%% file: main.tex
\documentclass[10pt,twocolumn,letterpaper]{article}

\usepackage{iccv}
\usepackage{times}
\usepackage{epsfig}
\usepackage{graphicx}
\usepackage{amsmath}
\usepackage{amssymb}
\usepackage{mmstyle}
\usepackage{marvosym}
\usepackage{placeins}
\usepackage{makecell} 
\usepackage{longtable}

\usepackage{subfigure}
\usepackage[font={small}]{caption}

\input{shortcuts.tex}

\usepackage[pagebackref=true,breaklinks=true,letterpaper=true,colorlinks,citecolor=blue,bookmarks=false]{hyperref}

\iccvfinalcopy 

\def\iccvPaperID{xxxx} 

\begin{document}

\title{V3Det: Vast Vocabulary Visual Detection Dataset}

\author{
Jiaqi Wang$^{* 1}$,
Pan Zhang$^{* 1}$,
Tao Chu$^{* 1}$,
Yuhang Cao$^{* 2}$, \\
Yujie Zhou$^{1}$,
Tong Wu$^{2}$,
Bin Wang$^{1}$,
Conghui He$^{1}$,
Dahua Lin$^{1,2,3}\textsuperscript{\Letter}$ \\
$^1$Shanghai AI Laboratory,
$^2$The Chinese University of Hong Kong, \\
$^3$Centre of Perceptual and Interactive Intelligence \\
\tt\small
\{wangjiaqi,zhangpan\}@pjlab.org.cn, dhlin@ie.cuhk.edu.hk
}


\input{sections/teaser.tex}
\input{sections/abstract.tex}
\input{sections/introduction.tex}
\input{sections/related_work.tex}

\input{sections/methods.tex}
\input{sections/experiments.tex}

\input{sections/conclusion.tex}

\clearpage
\setcounter{table}{0}
\setcounter{figure}{0}
\renewcommand{\thetable}{A\arabic{table}}
\renewcommand{\thefigure}{A\arabic{figure}}
\input{sections/supplementary}

\clearpage
{\small
\bibliographystyle{ieee_fullname}
\bibliography{egbib}
}

\end{document}

%% file: shortcuts.tex
\usepackage{array}
\usepackage{times}
\usepackage{epsfig}
\usepackage{graphicx}
\usepackage{float}
\usepackage{wrapfig}
\usepackage{amsmath,amssymb,amsthm}
\usepackage{algorithm,algorithmicx,algpseudocode}
\usepackage{bm,xspace}
\usepackage{comment}
\usepackage{multirow}
\usepackage{balance}
\usepackage{url}
\usepackage{booktabs}
\usepackage{etoolbox,siunitx}
\usepackage{calc}
\usepackage{pifont,hologo}
\usepackage{color}
\usepackage{adjustbox}
\usepackage[normalem]{ulem}  
\usepackage[table]{xcolor}
\usepackage{colortbl}
\usepackage{pgfplots}
\usepackage{nicefrac}

\definecolor{lightgreen}{HTML}{D8ECD1}
\definecolor{lightorange}{HTML}{FFE4C4}

\definecolor{blue}{HTML}{0055cc}
\definecolor{red}{HTML}{cc1100}
\definecolor{orange}{HTML}{cc7700}
\definecolor{gray}{HTML}{efefef}
\definecolor{darkgreen}{rgb}{0.13, 0.55, 0.13}
\definecolor{darkgray}{HTML}{757575}

\renewcommand{\eqref}[1]{Eq.~\ref{#1}}

\newcolumntype{x}[1]{>{\centering\arraybackslash}p{#1}}
\newcolumntype{y}[1]{>{\raggedright\arraybackslash}p{#1}}
\newcolumntype{z}[1]{>{\raggedleft\arraybackslash}p{#1}}
\newcommand{\tablestyle}[2]{\setlength{\tabcolsep}{#1}\renewcommand{\arraystretch}{#2}\centering\footnotesize}

\DeclareMathSymbol{@}{\mathord}{letters}{"3B}

\makeatletter
\DeclareRobustCommand\onedot{\futurelet\@let@token\@onedot}
\def\@onedot{\ifx\@let@token.\else.\null\fi\xspace}
\def\eg{e.g\onedot} 
\def\ie{i.e\onedot}

\newcommand*{\Rom}[1]{\expandafter\@slowromancap\romannumeral #1@}
\newcommand*{\rom}[1]{\expandafter\romannumeral #1}

\def\1{\bm{1}}

\DeclareMathAlphabet{\mathsfit}{\encodingdefault}{\sfdefault}{m}{sl}
\SetMathAlphabet{\mathsfit}{bold}{\encodingdefault}{\sfdefault}{bx}{n}

\def\ie{\emph{i.e.}}

%% file: sections/teaser.tex
\twocolumn[{
\renewcommand\twocolumn[1][]{#1}
\maketitle
\begin{center}
    \centering
    \vspace{-25pt}
    \includegraphics[width=1.0\linewidth]{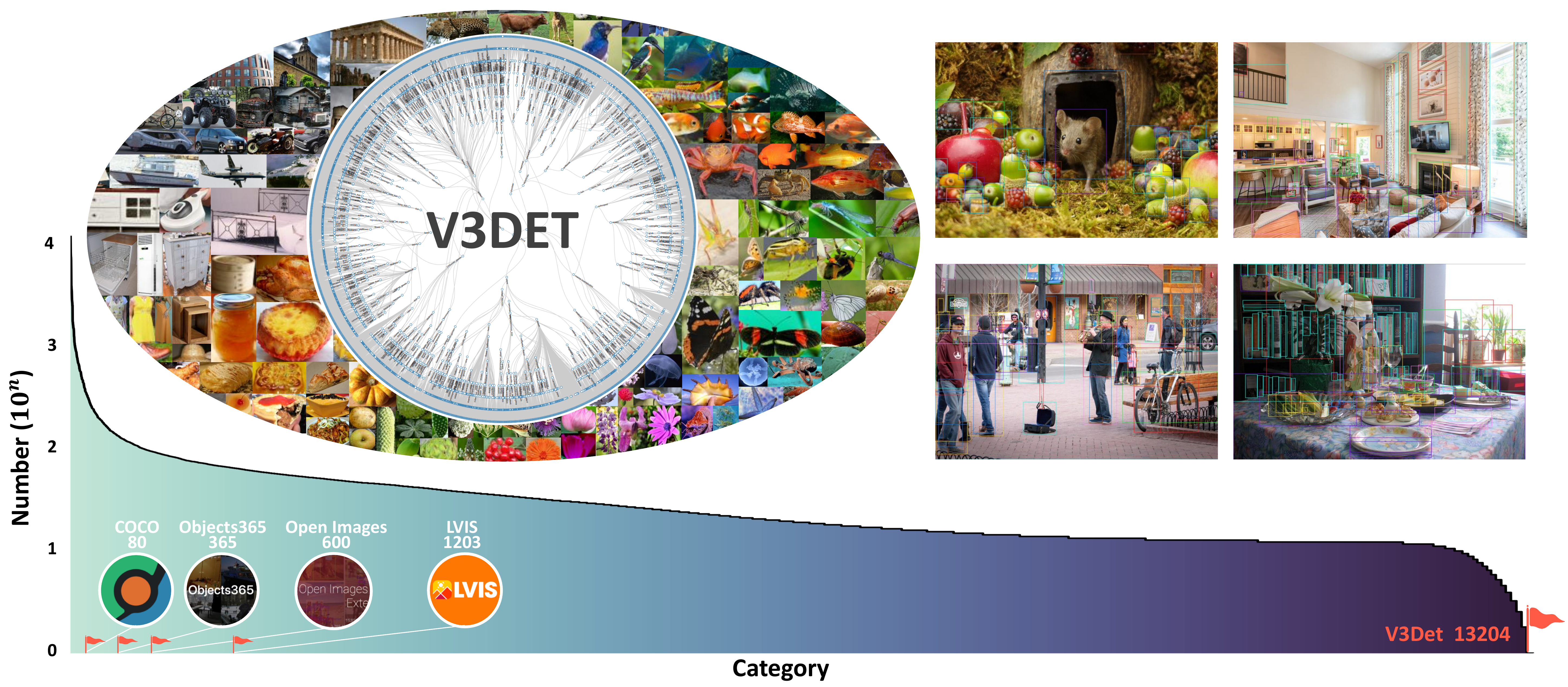}
    \setlength{\abovecaptionskip}{0mm}
    \captionof{figure}{\small
        \textbf{V3Det} is a \textbf{vast vocabulary visual detection} dataset with precisely annotated bounding boxes of 13,204 categories, significantly more diverse than existing visual detection datasets, \eg, COCO~\cite{lin2014coco}, Objects365~\cite{shao2019objects365}, OpenImages~\cite{kuznetsova2018open}, and LVIS~\cite{gupta2019lvis}. The hierarchical category tree, per-category object distributions, and annotated image samples of V3Det are shown in this figure. 
	}
	\label{fig:teaser}
    \vspace{0pt}
\end{center}
}]

%% file: sections/abstract.tex
\begin{abstract}

{\let\thefootnote\relax\footnotetext{\noindent * equal contribution.}}

Recent advances in detecting arbitrary objects in the real world are trained and evaluated on object detection datasets with a relatively restricted vocabulary. To facilitate the development of more general visual object detection, we propose \textbf{V3Det}, a vast vocabulary visual detection dataset with precisely annotated bounding boxes on massive images. V3Det has several appealing properties:   \textbf{1) Vast Vocabulary:} It contains bounding boxes of objects from 13,204 categories on real-world images, which is 10 times larger than the existing large vocabulary object detection dataset, \eg, LVIS. \textbf{2) Hierarchical Category Organization:} The vast vocabulary of V3Det is organized by a hierarchical category tree which annotates the inclusion relationship among categories, encouraging the exploration of category relationships in vast and open vocabulary object detection. \textbf{3) Rich Annotations:} V3Det comprises precisely annotated objects in 243k images and professional descriptions of each category written by human experts and a powerful chatbot. By offering a vast exploration space, V3Det enables extensive benchmarks on both vast and open vocabulary object detection, leading to new observations, practices, and insights for future research. It has the potential to serve as a cornerstone dataset for developing more general visual perception systems. V3Det is available at \url{https://v3det.openxlab.org.cn/}.

\end{abstract}

%% file: sections/introduction.tex
\section{Introduction}

Object detection~\cite{girshick2014rich,girshick2015fast,ren2015faster,cascade_rcnn,liu2016_ssd,lin2017_focal,carion2020end,PeizeSun2020SparseRE,wang2019region,Wang_2019_ICCV,Wang_2020_ECCV,wang2021carafeplus,Zhang_2023_CVPR} is the cornerstone of various real-world applications, \eg, autonomous driving, robotics, and augmented reality. Taking images as inputs, it localizes and classifies objects within a given vocabulary, where each detected object is denoted as a bounding box with a class label. 

Detecting arbitrary objects has been a long-standing goal in computer vision. Since the real world contains a vast variability of objects in countless classes, an ideal visual detection system should be able to detect objects from an extremely large set of classes and be readily applied to open-vocabulary categories. Therefore, it is imperative to equip the community with an object detection dataset with a vast vocabulary, as this can accelerate the exploration toward more general visual detection systems.

While there have been extensive efforts to create object detection datasets, they have only partially fulfilled the requirements for a comprehensive dataset with a vast vocabulary of categories. 
For example, as shown in Table~\ref{tab:dataset_comparison}, COCO~\cite{lin2014coco} is one of the most widely adopted datasets comprising 123k images with 80 categories. Objects365~\cite{shao2019objects365} and Open Images~\cite{kuznetsova2018open} are large-scale datasets aimed at pretraining, where Objects365 annotates 638k images with 365 categories, and Open Images contains 1,515k images with 600 categories. While these two datasets provide significant amounts of annotated objects, their limited category numbers still fall short in training class generalizable detectors. LVIS~\cite{gupta2019lvis} is a large vocabulary object detection and instance segmentation~\cite{huang2019msrcnn,chen2019hybrid,SOLO,tian2020conditional,chu2023buol,zhang2021prototypical,zhang2021robust,chen2019hybrid} dataset containing 120k images with 1,203 categories. 
Despite its success in recent advances in large / open vocabulary object detection, the vocabulary of LVIS (1,203 categories) is still insufficient to represent the vast diversity of classes in the real world. Therefore, a new object detection dataset with a significantly larger vocabulary would benefit the training and evaluation of visual detectors.

To facilitate future research on more general object detection, we introduce \textbf{V3Det}: \textbf{V}ast \textbf{V}ocabulary \textbf{V}isual Detection Dataset, which comprises a vast vocabulary of precisely annotated objects in real-world images. The proposed dataset has several appealing properties: 
\textbf{1) Vast Vocabulary:} It contains bounding boxes of objects from 13,204 categories on real-world images, which is 10 times larger than the existing large vocabulary object detection dataset, such as 1,203 categories in LVIS. 
\textbf{2) Hierarchical Category Organization: } The vast vocabulary of V3Det is organized by a hierarchical category tree that annotates the inclusion relationship among categories. This tree begins with `entity' category, which is the ancestor of all classes, containing both the categories of annotated objects and their parent categories as shown in Figure~\ref{fig:teaser}. This hierarchical tree enriches the vocabulary of V3Det, enabling the exploration of category relationships in open-world recognition.
\textbf{3) Rich Annotations: } V3Det comprises precisely annotated objects in 243k images. The large image collections ensure the diversity of objects and scenes. Besides category names, V3Det provides rich descriptions of each category written by human experts and a powerful chatbot, \ie, chatgpt, which can serve as language prompts for open-vocabulary object detection. 

Extensive benchmarks are conducted on V3Det to reveal insights into this vast vocabulary object detection dataset. Specifically, we perform extensive experiments on both classical close-set object detection and open-vocabulary object detection. 1) In close-set object detection, we evaluate various wide-adopted object detection frameworks. 
 We introduce the best practices of data samplers, optimizers, and classifier designs to achieve higher performance on V3Det.
2) In the open-vocabulary setting, we benchmark recent representative methods on V3Det, shedding light on the challengings of open-vocabulary object detections on a vast vocabulary of categories. Moreover, we show that pretraining on V3Det significantly boosts the class generalizability of detectors, demonstrating the solid benefits of a vast vocabulary dataset for open vocabulary algorithms. 

\begin{table}
\centering
\tablestyle{2pt}{1.1}
\begin{tabular}{c|ccccc}
\toprule
Dataset & Categories & Images & Boxes & Class Tree & Descriptions\\
\midrule
Pascal VOC~\cite{Everingham10} & 20 & 11.5k & 27k  & No & No \\
COCO~\cite{chen2015microsoft} & 80 & 123k & 896k  & No & No\\
Objects365~\cite{shao2019objects365} & 365 & 638k & 10,101k  & No & No\\
Open Images~\cite{kuznetsova2018open} & 600 & 1,515k & 14,815k  & Yes & No\\
LVIS~\cite{gupta2019lvis} & 1,203 & 120k & 1,525k  & No & No\\
ELEVATER~\cite{li2022elevater} & 314 & 132k & 938k  & No & Yes\\
BigDetection~\cite{cai2022bigdetection} & 600 & 3,480k & 35,960k  & No & No \\
\rowcolor{black!5}
V3Det & \textbf{13,029} & 215k & 1,540k  & Yes & Yes\\
\rowcolor{black!12}
V3Det (All) & \textbf{13,029} & 245k & 1,753k  & Yes & Yes\\
\bottomrule
\end{tabular}
\caption {Dataset statistics comparison between the V3Det and existing object detection benchmarks on the released annotations, including training and validation data. V3Det (All) includes all of training, validation, and testing data of V3Det.}
\vspace{-10pt}
\label{tab:dataset_comparison}
\end{table}

In summary, we introduce \textbf{V3Det}: \textbf{V}ast \textbf{V}ocabulary \textbf{V}isual Detection Dataset with 13,204 categories, hierarchically organized with a category tree. To the best of our knowledge, V3Det is the first object detection dataset with more than ten thousand categories. Rich annotations of detection bounding boxes in 243k images, along with professional category descriptions, ensure the diversity and versatility of the dataset. Taking advantage of the vast exploration space offered by V3Det, extensive benchmarks reveal new observations, practices, and insights for future research on vast and open vocabulary object detection. It has the potential to serve as a cornerstone dataset for developing more general visual perception systems.

%% file: sections/related_work.tex
\section{Related Work}
\noindent\textbf{Visual Detection Datasets.}
A large-scale dataset of broad categories is the key to building a robust visual detection system,  and the community has made many efforts. 
The early PASCAL VOC \cite{Everingham10} dataset contains 11.5K images of 20 categories.
The MS COCO \cite{lin2014coco} dataset introduces 123K images of 80 categories to depict complex everyday scenes of common objects.
Objects365 \cite{shao2019objects365}, Open Images \cite{kuznetsova2018open}, and BigDetection \cite{cai2022bigdetection} are
large-scale datasets for pretraining, where the largest BigDetection contains 3,480k images of 600 categories.
ELEVATER \cite{li2022elevater} is a multi-domain benchmark that comprises 35 datasets.
While these datasets provide significant amounts of annotated objects, their limited
category numbers still fall short in training class generalizable detectors.
LVIS \cite{gupta2019lvis} is the detection dataset of the largest vocabulary size yet, which has 1,203 categories and exhibits a long-tailed distribution.
Our V3Det dataset pursues a broader categories distribution that covers 13,204 classes, enabling training a more general detector.

\noindent\textbf{Object Detection.}
Object Detection aims to localize and classify objects, which has witnessed remarkable progress in recent years. 
The classical CNN-based detectors can be divided into two sets. 
Two-stage detectors\cite{girshick2014rich,girshick2015fast,ren2015faster,cascade_rcnn,zhou2021probabilistic} first predict a set of proposals and then refine them in the second stage. 
One-stage detectors \cite{lin2017_focal,Redmon_2017,liu2016_ssd}, which directly classify and regress the predefined anchor boxes, or densely searching geometric cues like points \cite{tian2019fcos}, centers \cite{duan2019centernet}, corners \cite{Law2018_CornerNet}.
Recent transformer-based detectors \cite{carion2020end,zhu2020deformable,zhang2022dino} remove hand-designed components like NMS and anchors and show superior performance than CNN-based ones.
In this work, we test various methods of different paradigms on our V3Det dataset.

\noindent\textbf{Open-Vocabulary Object Detection.}
Open-Vocabulary Object Detection (OVD) aims to detect novel classes not seen during training. 
To cope with the limitation of traditional detectors that detect only a predefined set of categories, OVD approaches usually adopt vision-language models \cite{radford2021learning,devlin2018bert} or utilize external data with weak annotations \cite{ILSVRC15,chen2015microsoft,sharma2018conceptual}.
Specifically, OV-RCNN \cite{zareian2021open} uses BERT \cite{devlin2018bert} to pretrain Faster R-CNN on a corpus of image-caption pairs. 
ViLD \cite{gu2021open} distills general knowledge from CLIP \cite{radford2021learning} into a two-stage detector.
GLIP \cite{li2022grounded} and MDETR \cite{kamath2021mdetr} convert detection as a grounding task with captions.
RegionCLIP \cite{zhong2022regionclip} enhances CLIP to match image regions and textual concepts better.
Detic \cite{zhou2022detecting} extends the classification capacity to twenty-thousand with image-level annotations from ImageNet-21K \cite{ILSVRC15}. 
Though these works achieve excellent performance on two benchmarks, MS COCO \cite{lin2014coco}, and LVIS \cite{gupta2019lvis}, we found by experiments that they can hardly apply in a scenario of vast categories.

%% file: sections/methods.tex

\begin{figure*}[t!]
\centering
\small
\setlength\tabcolsep{0pt}
\includegraphics[width=2.0\columnwidth]{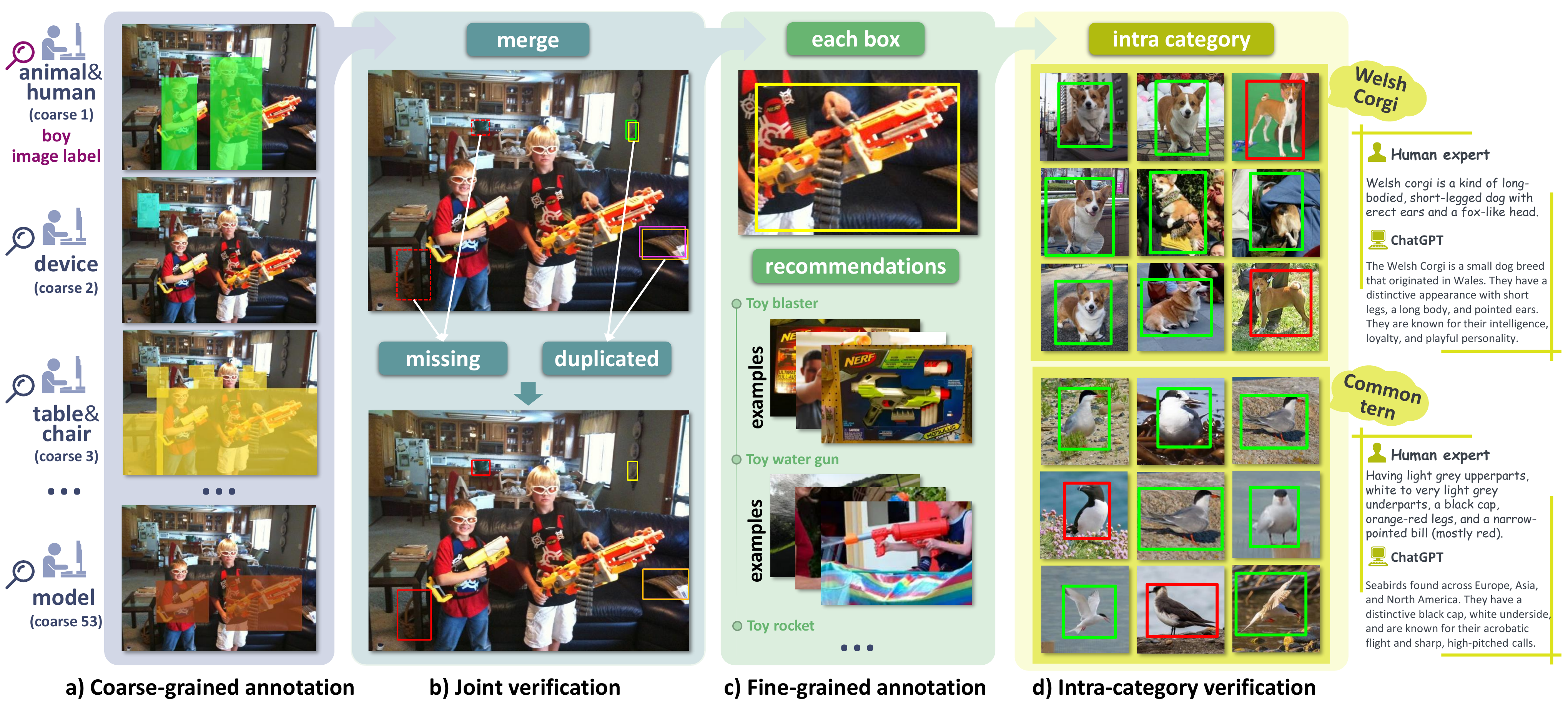}
\vspace{-5pt}
\caption{\textbf{The illustration of our annotation pipeline.} a) Each group is assigned to annotate one coarse-grained category and image-level categories belonging to the coarse-grained category. b) Annotations from all groups on the same image are merged and presented together for joint verification. c) For each box with coarse-grained annotation, some recommendations are given to the annotators for the fine-grained annotation. d) Objects annotated with the same fine-grained category are presented together for intra-category verification.}
\vspace{-15pt}
\label{fig:pipeline}
\end{figure*}

\section{V3Det Dataset} \label{sec:method}
This section presents a comprehensive exposition of the V3Det dataset, covering its data acquisition, annotation pipeline, and dataset analysis.

\subsection{Data Acquisition}
The construction of the V3Det dataset is based on the Bamboo~\cite{zhang2022bamboo}  classification dataset, where most categories and images utilized in the V3Det dataset are drawn from.

\noindent \textbf{Bamboo Classification Dataset.} The extensive Bamboo~\cite{zhang2022bamboo} classification dataset, containing 69 million annotations for image classification across 119,000 visual categories, serves as a rich source of potential images and categories for the V3Det dataset. Although the Bamboo classification dataset is a mega-scale dataset, not all of its categories are suitable for detection purposes, such as `sunny day' and `happiness'. Additionally, many classification images are less complex for the detection task. To address these issues, we developed a category construction and image selection process to obtain suitable categories and images for the V3Det dataset.

\noindent \textbf{Category Construction.} By carefully checking the available 119,000 visual categories, three researchers with ample expertise in object detection manually select visual concepts suitable for object detection. Each concept is equipped with a category name, a set of descriptions, and classification images. The description is sourced from human experts (\eg,WordNet, Wikidata) or generated by ChatGPT. The concepts unanimously chosen by all researchers will be incorporated into the V3Det dataset, resulting in 11,922 categories. In addition to selecting from the Bamboo classification dataset, we meticulously collect 1,282 object detection categories from web data. As a result, the V3Det dataset comprises a total of 13,204 categories.

\noindent \textbf{Image Selection.} For categories obtained from the Bamboo classification dataset, we adopt their corresponding images. For categories collected from web data, we crawl 120 images per category from Flickr \footnote{\url{https://www.flickr.com}} and manually remove any mismatched images. To filter the simple classification images, we employ a region proposal network (RPN)~\cite{ren2015faster} jointly trained on four detection datasets, namely COCO~\cite{lin2014coco}, LVIS~\cite{gupta2019lvis}, Objects365~\cite{shao2019objects365}, and OpenImages~\cite{kuznetsova2018open}, to extract region proposals for each image. Images with proposal counts exceeding 4,000 are removed to avoid overcrowding in the dataset. The remaining 20 images per category with the largest number of proposals are kept in the V3Det dataset.

\subsection{Annotation} 
The overall annotation pipeline of V3Det follows a coarse-to-fine strategy, as shown in Figure~\ref{fig:pipeline}. In the coarse-grained annotation and joint verification stages, objects belonging to coarse-grained categories and fine-grained image classification labels are annotated with bounding boxes. In the fine-grained annotation and intra-category verification stages, the coarse-grained category of each bounding box is further refined to the corresponding fine-grained category.

\noindent \textbf{Annotation Team.} To improve the accuracy of the annotation, we establish four annotation teams. One large Annotator Team performs all annotation steps in the entire annotation pipeline. Two Inspector Teams are assigned to review all the annotated images. Inspector Team \uppercase\expandafter{\romannumeral1} is responsible for reviewing all the annotations made by the Annotator Team. Once an annotation error is found in a data package (containing 100 images), the package will be returned for re-annotation. The data package approved by Inspector Team \uppercase\expandafter{\romannumeral1} will be further reviewed by Inspector Team \uppercase\expandafter{\romannumeral2}. 
Upon successful approval by the two Inspection Teams, one Examiner Team will finally examine the data packages. Any annotation errors will result in the data package being returned for re-annotation. The described entire process will be conducted in every stage of the annotation pipeline.

\noindent \textbf{Coarse-grained Annotation.} It is not feasible for each annotator to memorize and annotate all 13,204 categories. Therefore, we first divide the 13,204 categories into 53 coarse-grained categories, \eg, `animal and human', `vehicle', and `device'. The annotators are then trained to comprehend a single coarse-grained category and memorize the corresponding subcategories. Due to the similar appearance in each subcategory set, it is easier for annotators to complete the above task. To avoid misannotations, the annotators are divided into 53 groups, and each group annotates bounding boxes corresponding to a single coarse category across the entire dataset. For each group, numerous images may not contain any objects corresponding to the assigned coarse-grained category, leading to a large number of skipped images with no annotations. But extensive annotations are necessary to ensure the annotation accuracy.

When annotating coarse-grained bounding boxes, annotators also identify and annotate objects of image-level labels if it belongs to their assigned coarse category. As shown in Figure~\ref{fig:pipeline} a), the annotator of coarse group 1 (`animal and person') also annotates the bounding box of the image label (`boy'). Image-level classification labels are either from the Bamboo classification dataset or assigned after crawling images from the internet. Upon completion of the coarse-grained annotation stage, the resulting dataset comprises images annotated with coarse-grained bounding boxes and image-level fine-grained bounding boxes. 

\noindent \textbf{Joint Verification.} Although the `one annotator one coarse-grained category' strategy significantly reduces the difficulty of the annotations, there are still some annotation problems that need to be addressed. One is duplicated annotation. The other is missing annotation. 
Due to the mega-scale categories in the dataset, it is hard to ensure that coarse-grained categories are mutually exclusive. For instance, the category `truck' belongs to both the coarse-grained categories `vehicles' and `heavy machinery'. As a result, one annotator may annotate a bounding box with `vehicles' for a truck, while another annotator may annotate a box with `heavy machinery' for the same truck. Meanwhile, some fine-grained categories may not fit within a coarse-grained category well, which may lead to missed annotations from a small number of annotators.

To solve the aforementioned problems, a joint verification stage is proposed. As shown in Figure~\ref{fig:pipeline} b), the annotations from 53 annotation groups on each image are merged and forwarded to proficient annotators who have demonstrated exceptional skill in the coarse-grained annotation stage. The annotators are tasked with correcting any duplicated or missed annotations. The inspector and examiner teams review and examine this process to ensure the high quality of the joint verification stage.

\begin{figure*}[t]
    \center
    \small
    \setlength\tabcolsep{1pt}
    {
    \begin{tabular}{ccc}
        \includegraphics[width=0.7\columnwidth, trim=0cm 1cm 3cm 0cm, clip]{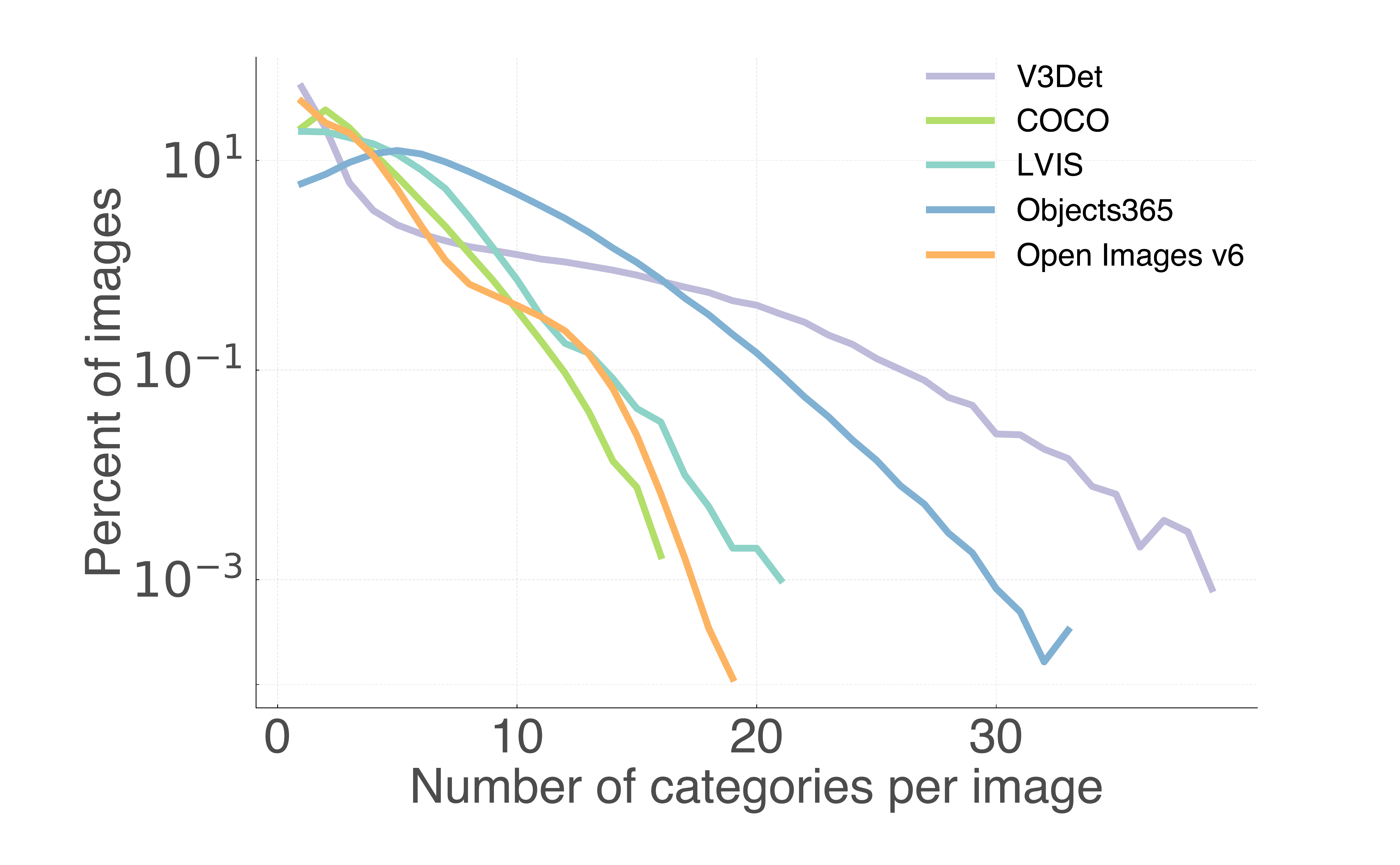} & 
        \includegraphics[width=0.6\columnwidth]{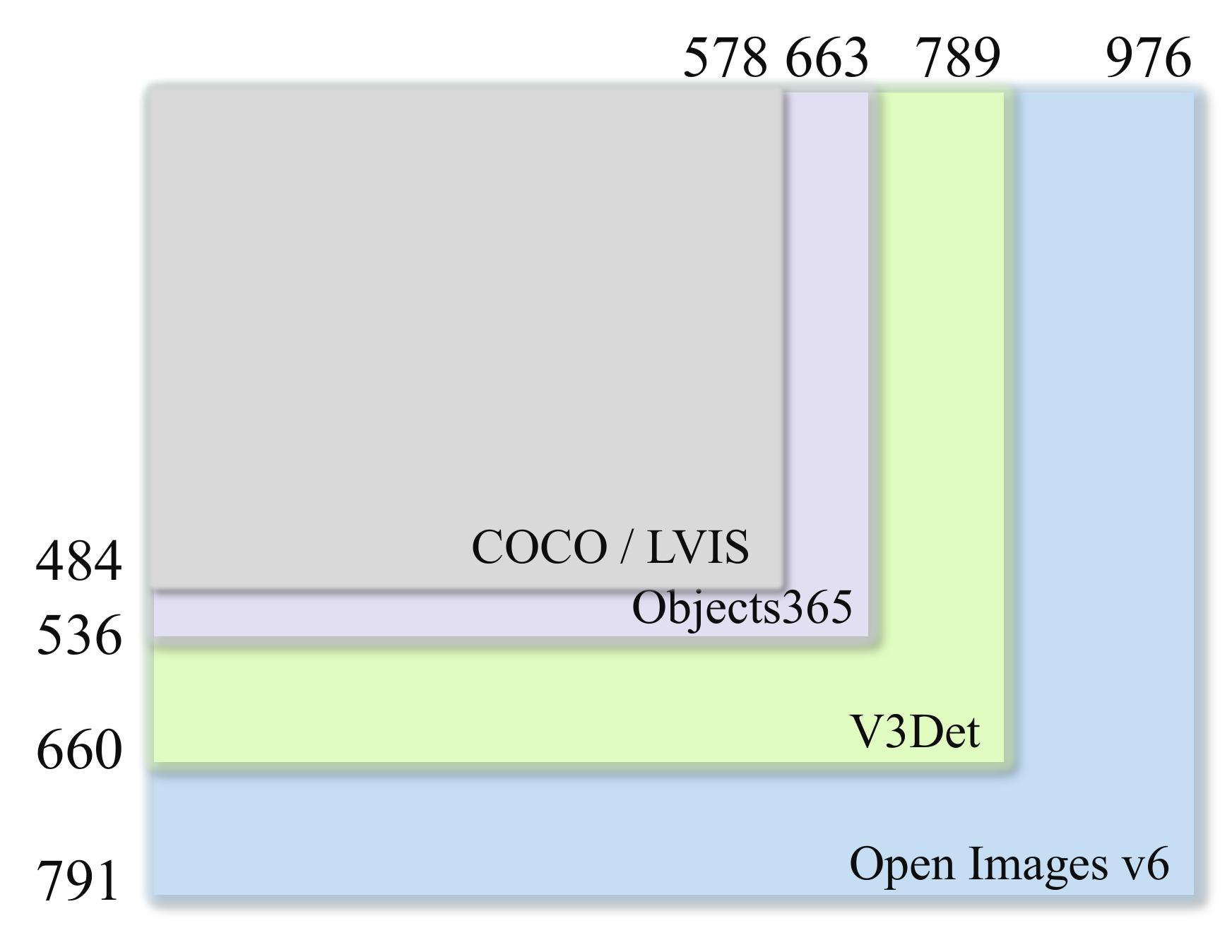} &
         \includegraphics[width=0.65\columnwidth, trim=3cm 1cm 5cm 0cm, clip]{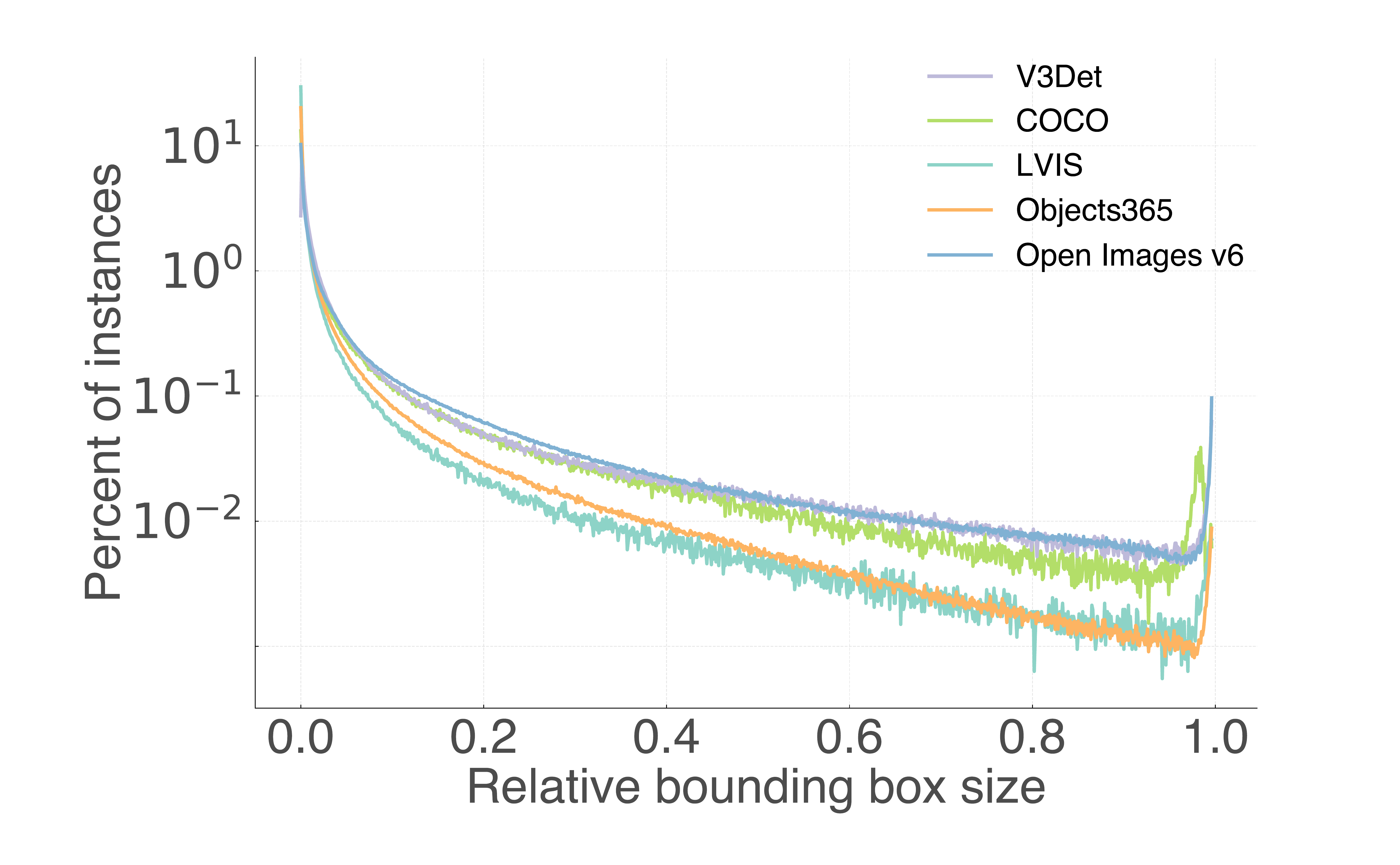} \\
        a) Distribution of category count per image & b) Mean resolution of each dataset & c) Distribution of relative bounding box size
    \end{tabular}
    }
    \vspace{-5pt}
\caption{\textbf{Dataset analysis.} a) The percentage of images containing more than 16 categories in our V3Det is greater than that in other dataset. b) The mean image resolution of our V3Det is higher than COCO, LVIS and Objects365. c) The object scale in our V3Det is comparable to that of the COCO and Open Images v6 datasets.}
\label{figure:dataset_statistics}
\vspace{-15pt}
\end{figure*}

\noindent \textbf{Fine-grained Annotation.} After completing the previous stages, all objects within the dataset are annotated by bounding boxes with coarse-grained categories or fine-grained image-level categories. This stage aims to refine the coarse-grained category of each bounding box to its corresponding fine-grained category. 

To alleviate the complexity of the coarse-to-fine annotation task, fine-grained categories are ranked by leveraging bounding boxes with image-level labels to assist the annotators. Specifically, the image patches are firstly cropped based on annotated bounding boxes, with a fixed padding size of 25 pixels. Then an image encoder of a large pre-trained visual language model, \ie, CLIP (ViT-L)~\cite{radford2021learning} is utilized to extract the discriminative features of the cropped patches. To obtain the representation for a fine-grained category $i$, we cluster all patch features of category $i$ with k-means~\cite{hartigan1979algorithm} to get $k$ prototypes $c^1_i, c^2_i,..., c^k_i$. Following CLUE~\cite{chen2003content}, the similarity $s^x_i$ between a coarse-grained image patch $x$ with a fine-grained category $i$ is defined as:
\begin{equation}
    s^x_i = \max_{j=1,2,...,k} \frac{e(x) \cdot c^j_i}{\lVert e(x) \rVert \lVert c^j_i \rVert}
\end{equation}
where $e(x)$ is the extracted feature of patch $x$, and $k$ is set to 3. Due to mega-scale categories, ranking all fine-grained categories solely based on the similarity score $s^x_i$ gives sub-optimal recommendations. Here we leverage the coarse-grained annotations. Specifically, for a coarse-grained category $I$, we construct a set of coarse-grained categories $\mathcal{I}_m$ that are more likely to be misannotated as category $I$. For a given cropped patch annotated with $I$, 
a traversal list is constructed with top-20 fine-grained categories belonging to coarse-grained category $I$, along with top-10 fine-grained categories whose coarse-grained categories is in $\mathcal{I}_m$. 

 During the fine-grained annotation stage, annotators are required to match a given cropped patch with all fine-grained categories in the traversal list. If none of the categories in the traversal list is correct, annotators are required to search in all fine-grained categories based on their own knowledge. The search platform provides relevant categories based on the search keyword as suggestions. As shown in Figure~\ref{fig:pipeline} c), the annotators choose the corresponding fine-grained category according to example images and class descriptions. 

After the above steps, there is still a small number of bounding boxes that are not successfully matched. To ensure the categories of these bounding boxes are not in the set of 13,204 categories, a small subset of the bounding boxes is sampled from each annotator, and a thorough check is conducted to confirm the results. An error rate greater than 2\%  results in re-annotation.

\noindent \textbf{Intra-category Verification.} After the fine-grained annotation stage, most annotation work of 13,204 categories has been completed. To enhance the accuracy of the annotations, a fine-grained intra-category verification stage is designed, where all the image patches cropped by bounding boxes with the same fine-grained category are displayed together, as in Figure~\ref{fig:pipeline} d). Annotators are tasked to verify if there are any misclassified objects or inconsistent annotations within the given fine-grained category with the help of example images and descriptions. The misclassified objects will be re-annotated in the fine-grained annotations stage, and the inconsistent annotations will be directly corrected by the annotator, following the same ``largest bounding box'' rule as the Objects365~\cite{shao2019objects365} dataset. This stage guarantees the accuracy of our V3Det dataset with 13,204 categories.

\subsection{Dataset Analysis} 
With the proposed annotation pipeline, approximately 243k images have been annotated with a total of 13,204 categories and 1.7M instances. The V3Det dataset exhibits the following main characteristics:

\noindent \textbf{Vast Vocabulary.} As indicated in Table~\ref{tab:dataset_comparison}. The number of the categories in the V3Det dataset is ten times greater than that of previous detection datasets. The extensive categories facilitate the development of the vast-vocabulary object detection. Moreover, since a real open-vocabulary object detector is also a vast-vocabulary object detector, the V3Det dataset, with its vast vocabulary, could provide a more robust benchmark to evaluate and support open-vocabulary object detection. Figure~\ref{figure:dataset_statistics} a) illustrates the full categories-per-image distribution, where the percentage of images containing more than 16 categories in V3Det is greater than that in other datasets, indicating the complexity and difficulty of the proposed V3Det dataset.

\begin{table*}[t]
	\begin{center}
            \tablestyle{12pt}{1.0}
\begin{tabular}{c|c|c|c|ccc}
\toprule
Framework                     & Method                                          &  Epochs             &  Backbone          & $AP$ & $AP_{50}$ & $AP_{75}$  \\ 
\midrule
\multirow{10}{*}{Two Stage / Cascade}    & \multirow{2}{*}{Faster R-CNN~\cite{ren2015faster}}                       & \multirow{2}{*}{24} & ResNet-50 &  20.1  &   28.2   &  22.9           \\
                                       &                                                     &                     & Swin-B    &  30.7  &  40.0    &  34.7             \\ \cmidrule{2-7}
                                       & \multirow{2}{*}{Faster R-CNN w/ Norm Linear Layer~\cite{wang2021seesaw}} & \multirow{2}{*}{24} & ResNet-50 &  25.4  &   32.9   &  28.1           \\
                                       &                                                     &                     & Swin-B    &  37.6  &   46.0   &  41.1            \\ \cmidrule{2-7}
                                       & \multirow{2}{*}{CenterNet2~\cite{zhou2021probabilistic}}                         & \multirow{2}{*}{24} & ResNet-50 &  28.3  &   33.7   &  30.4            \\
                                       &                                                     &                     & Swin-B    &  39.8  &   46.1   &   42.4         \\ \cmidrule{2-7}
                                       & \multirow{2}{*}{Cascade R-CNN~\cite{cai2019cascadercnn}}                      & \multirow{2}{*}{24} & ResNet-50 &  28.6  &  34.8    &   31.1            \\
                                       &                                                     &                     & Swin-B    &   40.2 &  47.3    &   43.0           \\ \cmidrule{2-7}
                                       & \multirow{2}{*}{Cascade R-CNN w/ Norm Linear Layer~\cite{wang2021seesaw}} & \multirow{2}{*}{24} & ResNet-50 &  31.6  &  37.3    &  33.7            \\
                                       &                                                     &                     & Swin-B    &  42.5  &   49.1   &   44.9           \\ \midrule
\multirow{6}{*}{Single Stage}          & \multirow{2}{*}{ATSS~\cite{chen2019hybrid}}                               & \multirow{2}{*}{24} & ResNet-50 &  4.4  &   5.3   &   4.6         \\
                                       &                                                     &                     & Swin-B    &  7.6  &   8.9   &   8.0       \\ \cmidrule{2-7}
                                       & \multirow{2}{*}{FCOS~\cite{tian2019fcos}}                               & \multirow{2}{*}{24} & ResNet-50 &  6.5  &  8.1    &   6.9            \\
                                       &                                                     &                     & Swin-B    &  15.0  &   17.9   &   16.0       \\ \cmidrule{2-7}
                                       & \multirow{2}{*}{FCOS w/ Norm Linear Layer~\cite{wang2021seesaw}}                               & \multirow{2}{*}{24} & ResNet-50 &  9.4 &  11.7   &   10.1         \\
                                       &                                                     &                     & Swin-B    &  21.0  &  24.8   &  22.3      \\ \midrule
\multirow{4}{*}{DETR Style}    & \multirow{2}{*}{Deformable DETR~\cite{zhu2020deformable}}                    & \multirow{2}{*}{50} & ResNet-50 &  34.4  &   39.9   &  36.4          \\
                                       &                                                     &                     & Swin-B  &  42.5 &   48.3  &   44.7    \\ \cmidrule{2-7}
                                       & \multirow{2}{*}{DINO~\cite{zhang2022dino}}                               & \multirow{2}{*}{24} & ResNet-50 &  33.5  &   37.7   &  35.0      \\
                                       &                                                     &                     & Swin-B    &  42.0  &   46.8   &  43.9          \\ \bottomrule
\end{tabular}
	\end{center}
    \vspace{-15pt}
 	\caption{
		\small{We report benchmark results of close-set vast vocabulary object detection setting on V3Det. AdamW~\cite{loshchilov2018decoupled} optimizer and repeat factor sampler~\cite{gupta2019lvis} are adopted for better performance in this table. We report $AP$, $AP_{50}$, and $AP_{75}$ for comparsions.
		}
	}\label{tab:close_res}
	\vspace{-15pt}
\end{table*}

\noindent \textbf{High Resolution.} As shown in Figure~\ref{figure:dataset_statistics} b), the higher image resolution of the V3Det, as compared to COCO, LVIS and Objects365,  provides greater visual details that can aid in object detection. Figure~\ref{figure:dataset_statistics} c) shows the relative size distribution of
bounding boxes. The object scale of V3Det is comparable to that of the COCO and Open Images v6.

\noindent \textbf{High Quality.} To validate the annotation quality of the V3Det dataset, three researchers with ample expertise in object detection are asked to annotate 100 images randomly sampled from V3Det dataset and 100 images randomly sampled from LVIS dataset. The results demonstrate that V3Det performs competitively with LVIS in terms of class-specific recall, achieving 90.9\% versus 91.2\%. In addition, V3Det outperforms LVIS in terms of mean 
average precision, achieving a higher rate of 86.5\% compared to 81.2\%.

%% file: sections/experiments.tex

\section{Benchmarks}

\subsection{Experimental Settings}\label{sec:exp_set}

\noindent
\textbf{Split Setup.} V3Det is divided into three splits, \ie, a \emph{train} split, a \emph{val} split, and a \emph{test} split. The \emph{train} split has 1,299,765 objects in 183,354 images, the \emph{val} split has 214,416 objects in 29,821 images, and the \emph{test} split has 212,437 objects in 29,863 images. We manually ensure at least one object of each category in the \emph{val} and \emph{test} split. The images and annotations of the \emph{train} and \emph{val} split will be publicly available. We will release the images of the \emph{test} split and set up a public test server to evaluate the uploaded results for fair comparisons. If not further specified, the reported performance is trained on \emph{train} split and evaluated on \emph{val }split.
We perform benchmarks on both close-set vast vocabulary object detection and open vocabulary object detection. 1) In the vast vocabulary setting, we take all the 13,204 categories for training and evaluation. 2) In the open vocabulary setting, we randomly sample 6,709 categories as the base classes $C_{base}$ and the remaining 6,495 categories as the novel classes $C_{novel}$.

\noindent
\textbf{Evaluation Protocol.} We follow the evaluation metrics of the COCO~\cite{chen2015microsoft} dataset to report the mean average precise (AP) on different IoU thresholds (\ie, 0.5 $\sim$ 0.95).

\noindent
\textbf{Implementation Details.} 1) In the vast vocabulary setting, we evaluate two-stage / cascaded detectors \cite{ren2015faster,cascade_rcnn,zhou2021probabilistic}, single-stage detectors \cite{lin2017_focal,tian2019fcos,zhang2020bridging}, and DETR-style detectors \cite{zhu2020deformable,zhang2022dino}. We take their standard implementations in mmdetection \cite{mmdetection}, Detectron2 \cite{wu2019detectron2}, and detrex \cite{ideacvr2022detrex}. All models are trained with AdamW~\cite{loshchilov2018decoupled} optimizer, multi-scale data augmentation, and repeat factor sampler~\cite{gupta2019lvis} as the default setting.
 2) In the open vocabulary object detection, we follow the officially released code and settings for training and inference. Please refer to supplementary materials for more implementation details.

\subsection{Experiments on Vast Vocabulary Settings}\label{sec:exp_close}

Compared to previous object detection datasets, V3Det contains a vast vocabulary of 13,204 categories. We first perform benchmarks on close-set vast vocabulary object detection to reveal insights and best practices on V3Det.

\noindent
\textbf{Main Results.} We evaluate detectors with two-stage, cascaded, single-stage, and DETR-style frameworks. Due to the limited resources, we can only test some representative detectors, but we believe the selected frameworks are sufficient to reveal important insights. As shown in Table~\ref{tab:close_res}, we report the performance of Faster R-CNN \cite{ren2015faster}, CenterNet2 \cite{zhou2021probabilistic}, and Cascade R-CNN \cite{cai2019cascadercnn} for two-stage / cascaded frameworks with ResNet-50~\cite{He_2016} and Swin-B~\cite{liu2021swin} backbones. Experimental results show that cascaded frameworks (\eg, Cascade R-CNN, CenterNet2) significantly outperform two-stage detectors (\eg, Faster R-CNN) with around 8\% on ResNet-50 and around 10\% on Swin-B. Single-stage detectors, \eg, FCOS~\cite{tian2019fcos} and ATSS \cite{zhang2020bridging} perform much inferior to Faster R-CNN and Cascade R-CNN on V3Det. The experimental results show that the cascaded refinement of regression and classification is crucial in V3Det. The current leading paradigm DETR-style detectors, \eg, Deformable DETR~\cite{zhu2020deformable} and DINO~\cite{zhang2022dino}, achieve strong performance on V3Det. For example, DINO~\cite{zhang2022dino} achieve higher performance than Cascade R-CNN~\cite{cai2019cascadercnn} with a large margin on ResNet-50 (33.5\% v.s. 28.6\%) and a clear margin on Swin-B (42.0\% v.s. 40.2\%).

Moreover, we evaluate the Norm Linear Layer~\cite{wang2021seesaw}, which is effective on large-vocabulary object detection dataset, LVIS. The Norm Linear Layer significantly improves the performance of Faster R-CNN, Cascade R-CNN, and FCOS with different backbones on the V3Det dataset.

\noindent
\textbf{Best Practices on V3Det Dataset.}
We explore the effectiveness of different data samplers, optimizers, and classifiers on V3Det. We take cascade R-CNN w/ ResNet-50 backbone with 12 epochs training as our baseline. In Table~\ref{tab:sampler_abl}, we compare random sampler, class balanced sampler~\cite{ouyang2016factors}, and repeat factor sampler~\cite{gupta2019lvis}. We observe that the repeat factor sampler and the class balanced sampler perform comparably, slightly better than the random sampler ($\sim$ 0.2\%). The results are consistent with the fact that V3Det dataset has a relatively balanced category distribution.
In Table~\ref{tab:opt_abl}, we perform an ablation study of AdamW optimizer and Norm Linear Layer. Both AdamW optimizer and Norm Linear Layer bring significant gains on V3Det. Therefore, in Table~\ref{tab:close_res}, we take the AdamW optimizer and repeat factor sampler as the default setting.

\begin{table}[t]
	\begin{center}
            \tablestyle{10pt}{1.2}
            \begin{tabular}{c|ccc}
            \toprule
             $\textbf{Data Sampler}$                   & $AP$ & $AP_{50}$ & $AP_{75}$ \\
            \midrule
            Random Sampler         &  25.7  &  31.7    &  28.1    \\
            Class Balanced Sampler~\cite{ouyang2016factors} &  25.9   &  31.8    &   28.3   \\
            Repeat Factor Sampler~\cite{gupta2019lvis}  &  25.9  &   32.0  &  28.4   \\ \bottomrule
            \end{tabular}
	\end{center}
        \vspace{-15pt}
 	\caption{
		\small{Study on different data samplers using Cascade R-CNN ResNet-50 with 12 epochs training as the baseline. AdamW optimizer is adopted in this table.
		}
	}\label{tab:sampler_abl}
	\vspace{-15pt}
\end{table}

\noindent
\textbf{Evaluate Large Foundation Model on V3Det Dataset.}
Recent advances in the foundation model dramatically boost the performance on existing object detection benchmarks, \eg, COCO and LVIS. Taking a recent foundation model EVA~\cite{fang2022eva} as an example, its ViT-G~\cite{dosovitskiy2020vit} version is first pre-trained on Objects365~\cite{shao2019objects365} dataset and then finetuned on the target dataset, achieving 64.2\% AP on COCO and 62.2\% AP on LVIS. Notably, LVIS is always considered a much more challenging dataset than COCO due to its large vocabulary and long-tailed distributions. However, EVA achieves comparable performance on both datasets. We explore whether such a powerful model can still achieve such a strong performance on V3Det. As in Table~\ref{tab:strong_res}, we report the performance of EVA w/ ViT-G on COCO, LVIS, and V3Det. To have a fair comparison, all training settings and model architecture are the same in experiments. Compared to 60+$\%$ AP on COCO and LVIS, there exists still wide space to improve the performance on V3Det. 

\begin{table}[t]
	\begin{center}
            \tablestyle{7pt}{1.2}
                \begin{tabular}{c|c|ccc}
                \toprule
                AdamW & Norm Linear Layer & $AP$ & $AP_{50}$ & $AP_{75}$ \\
                \midrule
                          &                   &   20.9 &  25.3    &   22.6   \\
                \checkmark          &                   &  25.9  &   31.8   &   28.3   \\
                          &     \checkmark              &  27.3  &   33.8   &   29.8   \\
                \checkmark          &    \checkmark               &  28.3  &  34.5    &  30.8   \\
                \bottomrule
                \end{tabular}
	\end{center}
        \vspace{-15pt}
	\caption{
		\small{Study on AdamW optimizer and Norm Linear Layer using Cascade R-CNN ResNet-50 with 12 epochs training baseline.
		}
	}\label{tab:opt_abl}
	\vspace{-15pt}
\end{table}

\begin{table}[t]
	\begin{center}
            \tablestyle{12pt}{1.2}
            \begin{tabular}{c|ccc}
            \toprule
             $\textbf{Dataset}$                   & $AP$ & $AP_{50}$ & $AP_{75}$ \\
            \midrule
            COCO~\cite{lin2014coco} &   64.2  &   81.9   &   70.6   \\
            LVIS~\cite{gupta2019lvis} &   62.2  &   76.2   &   65.4   \\
            V3Det  &  49.4  &  54.8  &  51.4   \\ \bottomrule
            \end{tabular}
	\end{center}
        \vspace{-15pt}
        \caption{
		\small{Comparison of the detection performance of a strong foundation model, \ie, EVA~\cite{fang2022eva} on COCO, LVIS, and V3Det.
		}
	}\label{tab:strong_res}
	\vspace{-15pt}
\end{table}

\noindent
\textbf{Performance Differences Between \emph{Val} and \emph{Test} Split.} We evaluate Cascade R-CNN~\cite{cai2019cascadercnn} and DINO~\cite{zhang2022dino} with 24 epochs and Swin-B backbone on \emph{val} and \emph{test} split of V3Det. The performance on \emph{test} is just slightly higher than \emph{val} split with 0.6\%$AP$ and 0.4\%$AP$ on the two methods, respectively, showing the consistency between the two splits.

\subsection{Experiments on Open Vocabulary Settings.}\label{sec:exp_open}
Detecting objects of arbitrary categories is a long-standing goal of object detection. In recent years, open vocabulary object detection has attracted more and more attention that aims to detect objects of any category after training on a close set of categories. In this section, we evaluate existing open vocabulary object detection (OVD) works on V3Det for more comprehensively measuring their category generalization ability.

\noindent
\textbf{Evaluate Pre-trained OVD models.} We first test various OVD methods via direct inference on V3Det \emph{val} split. As shown in Table~\ref{tab:direct_ovd}, we report the results on $AP$ of all categories, $bAP$ of base classes $C_{base}$, $nAP$ of novel classes on $C_{novel}$. Experimental results show that all OVD methods trained on existing object detection datasets, \eg, COCO, LVIS, and objects365, fail to perform well on V3Det. Among them, Detic achieves the highest performance of only 1\% $AP$. According to Table~\ref{tab:close_res}, fully supervised detectors can obtain much superior performance on V3Det. This observation shows that OVD methods still have a large exploration space to improve themselves.

\begin{table}[t]
	\begin{center}
            \tablestyle{0.5pt}{1.3}
            \begin{tabular}{c|c|c|cccc}
            \toprule
             $\textbf{Method}$    & Framework  & Training Data  & $AP$   & $bAP$ & $nAP$\\
            \midrule
            PBL~\cite{gao2021towards} & MRCNN-R50  & COCO,CCCap,VG,SBU & 0.24  &  0.22    &  0.26    \\
            VL-PLM~\cite{zhao2022exploiting} & FRCNN-R50  & COCO &  0.10 &  0.08   &   0.19   \\
            PromptDet~\cite{feng2022promptdet} & MRCNN-R50 & LVIS,LAION &  0.61 &    0.58   &   0.63   \\
            ViLD~\cite{du2022learning} & MRCNN-R50 & LVIS &  0.52 &    0.49   &   0.55   \\
            DetPro~\cite{du2022learning} & MRCNN-R50 & LVIS &  0.53 &    0.52   &   0.55   \\
            GLIP~\cite{li2022grounded} & DyHead\cite{Dynamic_2021_CVPR}-SWT & O365,GoldG,CC3M,SBU  & -  &  -    &   0.91   \\
            OC-OVD~\cite{Hanoona2022Bridging} & FRCNN-R50 & LVIS,ImageNet &  0.97 &   0.92  &  1.03    \\
            RegionClip~\cite{zhong2022regionclip} & FRCNN-R50 & LVIS & 0.72  &  0.71    &  0.72    \\
            Detic~\cite{zhou2022detecting}  & CenterNet2-R50 & LVIS,ImageNet & 1.00  &  0.95  &   1.04  \\
             \bottomrule
            \end{tabular}
	\end{center}
        \vspace{-15pt}
        \caption{
		\small{Evaluation results of directly testing pretrained open-vocabulary object detectors on V3Det. The framework and training data are also reported. $AP$, $bAP$, and $nAP$ indicates the performance on all classes, base classes $C_{base}$, and novel classes $C_{novel}$ of V3Det. GLIP is evaluated only on $nAP$ due to its high inference cost with a large class number.
        For framework, MRCNN, FRCNN, R-50, and SWT indicate Mask R-CNN, Faster R-CNN, ResNet-50, and Swin Transformer Tiny Model.
        For training data, CCCap, VG, SBU, O365, and CC3M indicate COCO Caption \cite{chen2015microsoft}, Visual Genome \cite{krishna2016visual}, SBU Caption \cite{ordonez2011im2text}, Objects365 \cite{shao2019objects365}, and Conceptual Caption 3M \cite{sharma2018conceptual}, respectively. GoldG indicates gold grounding data curated by MDETR \cite{kamath2021mdetr}. 
		}
	}\label{tab:direct_ovd}
	\vspace{-10pt}
\end{table}

\begin{table}[t]
	\begin{center}
            \tablestyle{3pt}{1.3}
            \begin{tabular}{c|c|c|ccc}
            \toprule
             $\textbf{Method}$    & Framework  & Training Data  & $AP$   & $bAP$ & $nAP$\\
            \midrule
            Detic~\cite{zhou2022detecting}  & Centernet2-R50 & V3Det-$C_{base}$   &  17.7 &  30.2  &  6.7  \\
            RegionClip~\cite{zhong2022regionclip} & FR-CNN-R50 & V3Det-$C_{base}$ & 12.6  &  22.1   &  3.1   \\
             \bottomrule
            \end{tabular}
	\end{center}
        \vspace{-15pt}
	\caption{
		\small{Evaluation of two representative open-vocabulary object detection works with training on $C_{base}$ in \emph{train} split of V3Det.
		}
	}\label{tab:train_ovd}
	\vspace{-10pt}
\end{table}

\begin{table}[t]
	\begin{center}
            \tablestyle{7pt}{1.3}
            \begin{tabular}{c|cc}
            \toprule
             $\textbf{Training Data}$     & LVIS-$nAP$   & Objects365-$AP$ \\
            \midrule
            LVIS$^*$~\cite{gupta2019lvis} & 7.0 &      1.6     \\
            LVIS$^*$ + ImageNet-22k$^*$~\cite{ILSVRC15} &   13.2    &     3.4      \\
            V3Det$^*$  & 14.5 &      4.4     \\
             \bottomrule
            \end{tabular}
	\end{center}
        \vspace{-15pt}
        \caption{
		\small{Comparsions of category generalization. Detic is adopted as the OVD framework. 1) LVIS$^*$: LVIS \emph{train} split without categories of LVIS-rare and objects365, 2) LVIS$^*$ + ImageNet-22k$^*$: LVIS \emph{train} split without categories of LVIS-rare and objects365 + ImageNet-22k images with class labels in LVIS and objects365, 3) V3Det$^*$: V3Det \emph{train} split without categories of LVIS-rare and objects365. LVIS-$nAP$ and Objects365-$AP$ indicate the performance of LVIS rare classes and all Objects365 classes.
		}
	}\label{tab:general_ovd}
	\vspace{-10pt}
\end{table}

\noindent
\textbf{Evaluate OVD methods trained on V3Det.} We further explore the effectiveness of current OVD methods after training on V3Det. Specifically, we train two well-known OVD methods, \ie, Detic~\cite{zhou2022detecting} and RegionCLIP~\cite{zhong2022regionclip} on base classes $C_{base}$ in \emph{train} split of V3Det, and test them on both $C_{base}$ and $C_{novel}$ in \emph{val} split of V3Det. As shown in Table~\ref{tab:train_ovd}, after training on $C_{base}$ of V3Det, both methods gain considerable improvements on both $bAP$ and $nAP$. Notably, $nAP$ of Detic and RegionCLIP are significantly boosted from 1.04\% to 6.7\% and 0.72\% to 3.1\%, respectively. Although the performance is still far from that of supervised models, these experimental results confirm that a vast vocabulary dataset will lead to stronger category generalization ability.

\noindent
\textbf{Category Generalization of V3Det.} To further verify the effectiveness of V3Det in training open vocabulary object detectors, we evaluate Detic trained with V3Det on LVIS-rare and objects365 categories. We form a new subset of categories $C_{v3det}^* = C_{v3det} \setminus ( C_{lvis\_rare} \cup C_{objects365})$ which remove LVIS rare classes and Objects365 classes from the vocabulary of V3Det, and $C_{lvis}^* = C_{lvis} \setminus ( C_{lvis\_rare} \cup C_{objects365})$ which remove LVIS rare classes and Objects365 classes from the vocabulary of LVIS. After learning from $C_{v3det}^*$ and $C_{lvis}^*$, we report the performance of LVIS-$nAP$ of LVIS rare classes and Objects365-$AP$ of objects365 on the \emph{val} split of the two datasets. As shown in Table~\ref{tab:general_ovd}, Detic trained on $C_{v3det}^*$ significantly outperforms the one trained on $C_{lvis}^*$, even on rare classes of LVIS itself. Moreover, we also try to leverage ImageNet images with class labels in LVIS and objects365, which means `LVIS$^*$ + ImageNet-22k$^*$' setting contains target categories, \ie, LVIS-rare and objects365. Surprisingly, V3Det still performs better, showing its capability to serve as a dataset for further research on open vocabulary object detection.

%% file: sections/conclusion.tex
\section{Conclusion}
We introduce \textbf{V3Det}, a \textbf{v}ast \textbf{v}ocabulary \textbf{v}isual detection dataset with precisely annotated bounding boxes on massive real-world images. V3Det comprises 243k images in 13,204 categories. The vast vocabulary is organized by a hierarchical category tree. Professional descriptions of each category written by human experts and a powerful chatbot are available. Extensive benchmarks are performed on both vast and open vocabulary object detection, leading to new observations, practices, and insights for future research. 

\noindent\textbf{Acknowledgement.}
This project is funded by Shanghai AI Laboratory (P23KS00010, 2022ZD0160201), the Centre for Perceptual and Interactive Intelligence (CPIl) Ltd under the Innovation and Technology Commission (ITC)'s InnoHK, and CUHK Interdisciplinary AI Research Institute.

%% file: sections/supplementary.tex

{\let\thefootnote\relax\footnotetext{\noindent * equal contribution.}}

\appendix
In the supplementary materials, we introduce the license of V3Det dataset in Appendix~\ref{app:license}, more implementation details in Appendix~\ref{app:details}, and more experimental results in Appendix~\ref{app:exp}. We show a more detailed visualization of the hierarchy category organization of V3Det in Appendix~\ref{app:tree}, the list of coarse categories which is used during annotation process in Appendix~\ref{app:coarselist}, some examples of category descriptions written by human experts and a powerful chatbot, \ie, chatgpt\footnote{\url{https://openai.com/blog/chatgpt}} in Appendix~\ref{app:desc}, and more visualizations of the V3Det dataset in Appendix~\ref{app:vis}.

\section{V3Det Dataset License and Download.} \label{app:license}

\noindent \textbf{V3Det Images.}
Around 90\% of the images in V3Det were selected from the Bamboo Dataset~\cite{zhang2022bamboo}, sourced from the Flickr website. The remaining 10\% were directly crawled from the Flickr. We do not own the copyright of the images. Use of the images must abide by the Flickr Terms of Use\footnote{\url{https://www.flickr.com/creativecommons/}}. We only provide lists of image URLs without redistribution. 

\noindent \textbf{V3Det Annotations.} The V3Det annotations, the category relationship tree, and related tools are licensed under a Creative Commons Attribution 4.0 License~\footnote{\url{https://creativecommons.org/licenses/by/4.0/}}.

\noindent \textbf{V3Det Download.}
The metafile of image URLs, a script to easily download images, annotations, the category relationship tree, and other related tools are available at \url{https://v3det.openxlab.org.cn/}.

\section{Implementation Details.} \label{app:details}
\subsection{Vast Vocabulary Object Detection.}
\noindent \textbf{Two Stage and Cascaded Detectors.} We test three representative two-stage or cascaded detectors with ResNet-50 \cite{He_2016} and Swin-B \cite{liu2021swin} backbones, \ie, Faster R-CNN~\cite{ren2015faster}, Cascade R-CNN~\cite{cai2019cascadercnn}, and CenterNet2~\cite{zhou2021probabilistic}. For Faster R-CNN and Cascade R-CNN, we adopt their standard implementations in mmdetection \cite{mmdetection}. For CenterNet2, we adopt its official implementation based on Detectron2 \cite{wu2019detectron2}. To have a fair comparison, FPN~\cite{lin2017_fpn}, AdamW optimizer\cite{loshchilov2018decoupled}, multi-scale augmentation, and repeat factor sampler~\cite{gupta2019lvis} training are adopted in experiments. Specifically, we use AdamW optimizer \cite{kingma2014adam} with learning rate of $10^{-4}$, $\beta_1$=0.9, $\beta_2$=0.999, and batchsize of 32. Following the best practices on LVIS~\cite{gupta2019lvis} dataset, we use multi-scale augmentation training by randomly selecting the shorter side of the image from (640, 672, 704, 736, 768, 800) pixels and the longer size of the image less than 1333 pixels. Following LVIS~\cite{gupta2019lvis}, we set the repeat factor to $10^{-3}$ in the repeat factor sampler~\cite{gupta2019lvis}.
The detectors are trained with 24 epochs, where the learning rate is decreased by 10$\times$ at 16 and 22 epochs. 

\noindent \textbf{Single Stage Detectors.} We adopt the standard implementation of ATSS\cite{chen2019hybrid} and FCOS~\cite{tian2019fcos} in mmdetection \cite{mmdetection} as representative single stage detectors. Other settings follow the experiments on two stage and cascaded detectors.

\noindent \textbf{DETR Style Detectors.} 
We test two DETR-style detectors, Deformable DETR \cite{zhu2020deformable}, and DINO \cite{zhang2022dino} with two different backbones, ResNet-50 \cite{He_2016}, and Swin-B \cite{liu2021swin}. We implement Deformable DETR with mmdetection \cite{mmdetection} and train it with batch size 32 for 50 epochs with the learning rate of $2 \times 10^{-4}$.
We implement DINO with detrex \cite{ideacvr2022detrex} and train it with batch size 16 for 24 epochs with the learning rate of $10^{-4}$.
Both models are trained with AdamW optimizer \cite{kingma2014adam} with $\beta_1$=0.9, $\beta_2$=0.999 and repeat factor sampler with factor of $10^{-3}$.
We follow the default data augmentation that is adopted in the corresponding framework.

\subsection{Open Vocabulary Object Detection (OVD).}
In this section, we provide the implementation details of training Detic \cite{zhou2022detecting} and RegionCLIP \cite{zhong2022regionclip} on V3Det.

For Detic, we follow the original paper that uses CenterNet2 \cite{zhou2021probabilistic} with ResNet-50 \cite{He_2016}.
We adopt large-scale jittering~\cite{Ghiasi_2021_CVPR} with an input resolution of 640x640. 
The batch size is 64, and the learning rate is $2 \times 10^{-4}$ with the AdamW optimizer.
For results in main text Table 6, we first train a model on base classes $C_{base}$ of V3Det for 90k iterations. We then create a subset of ImageNet-21K, named IN-V3Det, which contains 4197 overlapped classes with V3Det. We then do a multi-dataset finetuning on V3Det and IN-V3Det for 90k iterations. 
For results in main text Table 8, Detic of V3Det$^*$ is only trained on V3Det$^*$ without ImageNet images for 180k iterations.

For RegionCLIP, following the original paper, we utilize Faster R-CNN~\cite{ren2015faster} with ResNet-50~\cite{He_2016} backbone that is finetuned by CLIP-guided region-text alignment. Initially, the offline Region Proposal Network (RPN)~\cite{ren2015faster} is trained on the base categories for 180k iterations. Subsequently, the finetune stage is applied to transfer the learned knowledge of region-text alignment to open-vocabulary detection for 180k iterations. During training, a batch size of 16 images and an initial learning rate of $2 \times 10^{-3}$ are utilized. 

\section{More Experimental Results.} \label{app:exp}
\noindent \textbf{V3Det for pretraining.} V3Det aims to be a benchmark for vast-vocabulary and open-vocabulary object detection. Interestingly, our findings indicate that V3Det can additionally be employed effectively as a pretraining dataset. We conducted comparative analyses of diverse strategies for pretraining a CenterNet2 \cite{zhou2021probabilistic} with R-50 on datasets V3Det and Objects365. Subsequent finetuning was performed on the LVIS dataset to assess the effectiveness of the initial pretraining. Table~\ref{tab:pretrain} shows that V3Det and Objects365 exhibit comparable performance and can complement each other for pretraining. 

 \begin{table}[t]
	\begin{center}
            \tablestyle{12pt}{1.0}
\begin{tabular}{c|c|c|c}
\toprule
Pretrain (1st) & Pretrain (2st) & Finetune & AP \\
\midrule
- & -  & LVIS & 35.1 \\
V3Det & -  & LVIS & 36.2 \\
Objects365 & -  & LVIS & 36.9 \\
V3Det & Objects365 & LVIS & 37.3 \\
Objects365 & V3Det  & LVIS & 37.7 \\
\bottomrule
\end{tabular}
\end{center}
\vspace{-15pt}
 	\caption{
		\small{Comparisons of different strategies when pretraining a CenterNet2 using R-50 backbone and Norm Linear Layer on V3Det and Objects365, followed by finetuning on LVIS. For each stage, the training schedule is 90k iterations of batch size 64.
		}
	}\label{tab:pretrain}
	\vspace{-10pt}
\end{table}

\noindent \textbf{Ablation on Norm Linear Layer.} In the experiments of Table 2 and Table 4 in the main text, we show the effectiveness of Norm Linear Layer~\cite{wang2021seesaw}, $z = \tau{\widetilde{\cW}}^T \widetilde{x} + b$, where $\widetilde{\cW}_{i}=\frac{\cW_{i}}{\left\|\cW_{i}\right\|_2}, i \in C$, $\widetilde{x} = \frac{x}{\left\|x\right\|_2}$. As shown in Table~\ref{tab:tau}, we explore its temperature factor $\tau$, and find that the best performance is achieved when $\tau$ is 50. Therefore, we adopt $\tau$ of 50 as the default setting in experiments.

\begin{table}[t]
	\begin{center}
            \tablestyle{12pt}{1.0}
\begin{tabular}{c|c|c|c}
\toprule
$\tau$ & $AP$ & $AP_{50}$ & $AP_{75}$ \\
\midrule
30 & 27.4  & 32.9 & 29.7 \\
40  & 27.8 & 34.0 & 30.3 \\
\rowcolor{black!8}
50  & \textbf{28.3} & \textbf{34.5} & \textbf{30.8} \\
60  & 27.8 & 33.9 & 30.3 \\
\bottomrule
\end{tabular}
\end{center}
\vspace{-15pt}
 	\caption{
		\small{Comparisons of different temperature factors $\tau$ in Norm Linear Layer. Cascade R-CNN ResNet-50 trained for 12 epochs with the AdamW optimizer is the framework in this table.
		}
	}\label{tab:tau}
	\vspace{-10pt}
\end{table}

\begin{table}[t]
  \vspace{5pt}
  \begin{center}
  \tablestyle{8pt}{1.0}
  \begin{tabular}{c|c|c|c}
    \toprule
    Stage & AP & AP50 & AP75 \\
    \midrule
    Faster R-CNN~\cite{ren2015faster} & 21.2 & 29.5 & 24.1 \\
    Cascade R-CNN stage 1 & 22.7 & 31.1 & 25.8 \\
    Cascade R-CNN stage 2 & 27.0 & 33.7 & 29.7 \\
    Cascade R-CNN all stage& \textbf{28.3} & \textbf{34.5} & \textbf{30.8} \\
    \bottomrule
  \end{tabular}
  \end{center}
  \vspace{-15pt}
  \caption{\small{Performance of Cascade R-CNN at different stages.}}
  \label{tab:cascade_stage_ap}
\vspace{-10pt}
\end{table}

\noindent \textbf{Different Stages in Cascaded R-CNN.} As in Table ~\ref{tab:cascade_stage_ap}, we show the performance of Cascade R-CNN \cite{cai2019cascadercnn} ResNet-50 at different stages. The AP increases as the stage grows, demonstrating the effectiveness of cascading refinement cascade. On the other hand, the AP of Cascade R-CNN at stage 1 is 22.7, which is still higher than the AP of Faster R-CNN \cite{ren2015faster}, which is 21.2, indicating the structure of stage cascade is beneficial to the model optimization. 

\begin{table}[t]
  \vspace{5pt}
  \begin{center}
  \tablestyle{1pt}{1.0}
  \begin{tabular}{c|c|c|c|c}
    \toprule
    Method & Split & $AP$ & $AP_{50}$ & $AP_{75}$ \\
    \midrule
    \multirow{2}{*}{Cascade R-CNN~\cite{cai2019cascadercnn} + Norm Linear Layer~\cite{wang2021seesaw}} & \emph{val} & 42.5 & 49.1 & 44.9 \\
                                                       & \emph{test} & 43.1 & 49.7 & 45.6 \\
    \midrule
    \multirow{2}{*}{DINO~\cite{zhang2022dino}} & \emph{val} & 42.0 & 46.8 & 43.9 \\
                          & \emph{test} & 42.4 & 47.2 & 44.3  \\
    \bottomrule
  \end{tabular}
  \end{center}
  \vspace{-15pt}
  \caption{\small{Performance differences between \emph{val} and \emph{test} split.}}
  \label{tab:ap_val_test}
\vspace{-10pt}
\end{table}

\noindent
\textbf{Performance Differences Between \emph{Val} and \emph{Test} Split.} In Table~\ref{tab:ap_val_test}, we evaluate Cascade R-CNN~\cite{cai2019cascadercnn} + Norm Linear Layer~\cite{wang2021seesaw} and DINO~\cite{zhang2022dino} with 24 epochs, AdamW optimizer and Swin-B backbone on \emph{val} and \emph{test} split of V3Det, showing the annotation consistency of the two splits.

\begin{table}[t]
    \vspace{5pt}
    \begin{center}
    \tablestyle{12pt}{1.0}
    \begin{tabular}{c|c|c|c}
      \toprule
      Dataset & AP & CLS ER & LOC ER \\
      \midrule
      LVIS~\cite{gupta2019lvis}  & \textbf{62.2} & 7.69  & \textbf{4.78} \\
      V3Det & 49.4 & \textbf{25.32} &  0.75 \\
      \bottomrule
    \end{tabular}
    \end{center}
    \vspace{-15pt}
    \caption{\small{Comparison of classification error (CLS ER) and localization error (LOC ER) of EVA \cite{fang2022eva} on LVIS \cite{gupta2019lvis} and V3Det.}}
    \label{tab:error_type}
\vspace{-10pt}
\end{table}
  
\noindent \textbf{EVA Error Analysis.} Table 5 in the main text shows the AP of EVA \cite{fang2022eva} on V3Det is 49.4, which is 12.8 lower than the AP on LVIS \cite{gupta2019lvis}, which is 62.2. In this section, we explore the error source of the AP difference. Table \ref{tab:error_type} compares the classification and localization errors of EVA on LVIS and V3Det, computed by TIDE \cite{tide-eccv2020}. 
Classification error indicates localized correctly (IoU $>$ 0.5) but classified incorrectly; Localization error indicates classified correctly but localized incorrectly (IoU $<$ 0.5).
We can see the localization error of V3Det (0.75) is lower than LVIS (4.78), but the classification error of V3Det (25.32) is much higher than LVIS (7.69). This confirms V3Det exposes a more challenging vast vocabulary classification problem than LVIS, leading to a broader exploration space.

\noindent \textbf{Hierarchy Open Vocabulary Test.} To comprehensively evaluate the open-vocabulary detectors, we propose a hierarchy open vocabulary test to evaluate the hierarchy capability of the detector, which is built upon our hierarchy category organization. Firstly, we employ two distinct methodologies for assessing the Average Precision metric of non-leaf nodes. One is Vocabulary based Non-leaf Average Precision $AP^{v}$, defined as  
\begin{equation}
\begin{aligned}
  AP^{v} = \frac{1}{N}\sum_{i \in \{\text{non-leaf nodes}\}}AP(P_i, f_i(Y)), \quad\quad\quad \\
  f_i(Y) = \{y | y \in Y \text{ and } y \in \text{descendants of node }i \},
\end{aligned}
\end{equation}
where $N$ is total number of non-leaf nodes in the hierarchy tree. $P_i$ is the predicted boxes of non-leaf node $i$. $Y$ is the ground truth of all leaf nodes. $f_i(Y)$ is all ground truth of the descendants of node $i$. $AP(\alpha, \beta)$ is the Average Precision between boxes set $\alpha$ and $\beta$. The other is Hierarchy based Non-leaf Average Precision $AP^{h}$, defined as 
\begin{equation}
\begin{aligned}
  AP^{h} = \frac{1}{N}\sum_{i \in \{\text{non-leaf nodes}\}}AP(f_i(P), f_i(Y)), \quad\quad \\
  f_i(P) = \{p | p \in P \text{ and } p \in \text{descendants of node }i \},
\end{aligned}
\end{equation}
where $P$ is the predicted boxes of all leaf nodes. $f_i(P)$ is the merged predictions of the descendants belonging to node $i$ processed by NMS with IoU threshold of 0.5.

Based on $AP^{v}$ and $AP^{h}$, we design Hierarchy Score, dubbed as H-score, which is defined as
\begin{equation}
     \text{H-score} = AP^{v} / (AP^{h} + \epsilon) \times 100,
\end{equation}
where $\epsilon$ is set to 1e-6. A higher value of the H-score indicates a stronger hierarchical capability of the detector. Table~\ref{tab:HOW} provides the results of the hierarchy open vocabulary test. Although the $AP^{v}$ and $AP^{h}$ of RegionCLIP are lower than that of Detic, the elevated H-score of RegionCLIP in comparison to Detic suggests a superior hierarchical capability, owing to the robustness of the leveraging CLIP\cite{radford2021learning} parameters rather than merely extracted text embeddings.

\begin{table}[t]
	\begin{center}
            \tablestyle{12pt}{1.0}
\begin{tabular}{c|c|c|c}
\toprule
Method & $AP^{v}$ & $AP^{h}$ & H-score \\
\midrule
Detic~\cite{zhou2022detecting} & 7.25 & 16.90 & 42.90 \\
RegionCLIP~\cite{zhong2022regionclip}  & 5.58 & 11.70 & 47.69 \\
\bottomrule
\end{tabular}
\end{center}
\vspace{-15pt}
 	\caption{
		\small{The results of the hierarchy open vocabulary test.
		}
	}\label{tab:HOW}
	\vspace{-15pt}
\end{table}

\section{Hierarchy Category Organization.} \label{app:tree}

\begin{figure*}[t]
    \center
    \small
    \setlength\tabcolsep{1pt}
    {
    \begin{tabular}{c}
        \includegraphics[width=\linewidth]{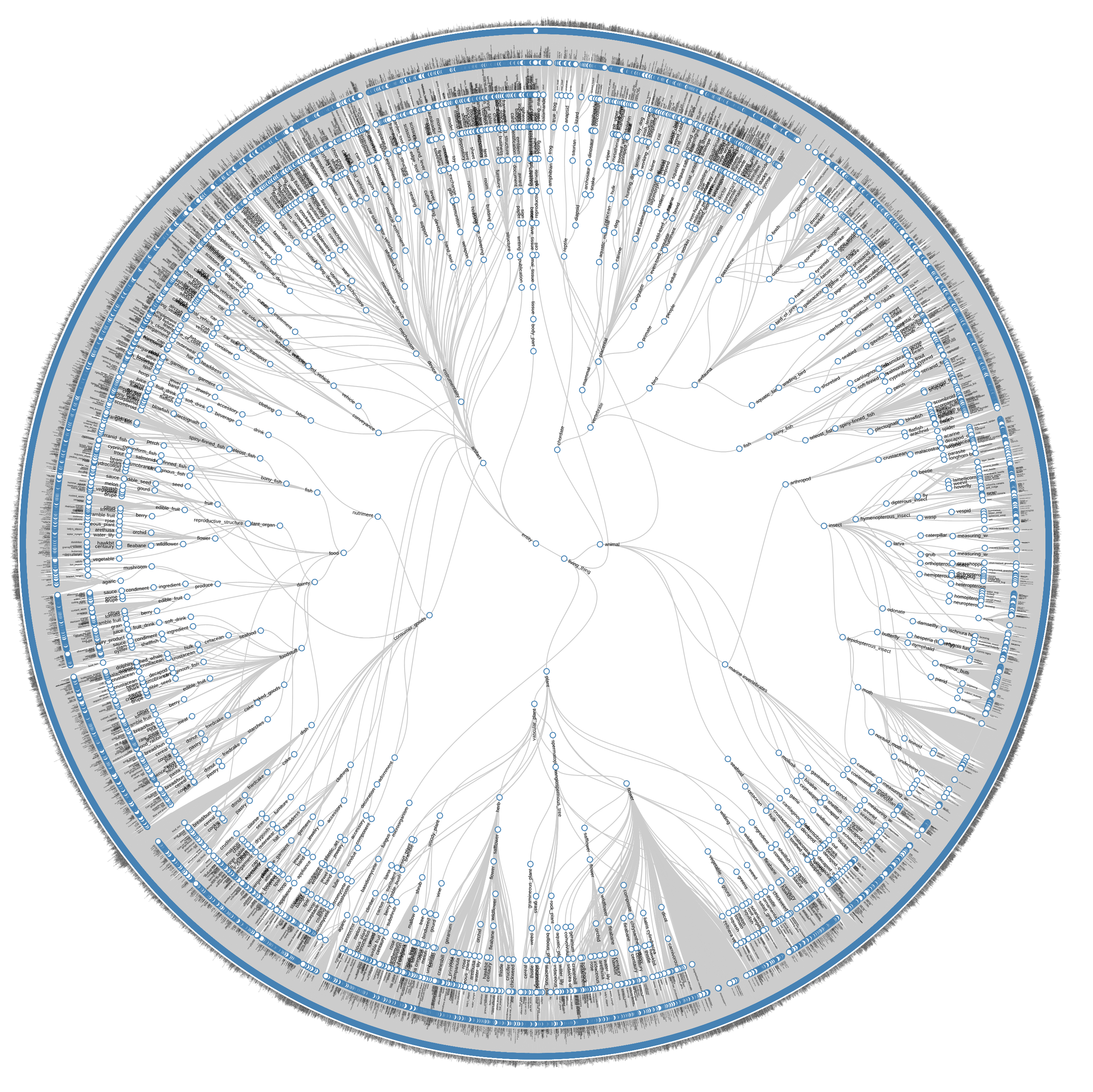}
    \end{tabular}
    }
    \vspace{-10pt}
\caption{\textbf{A Detailed Visualization of Hierarchy Category Organization in V3Det.} }
\label{figure:hierarchy}
\vspace{-10pt}
\end{figure*}

A more detailed hierarchy category organization of V3Det is shown in Figure~\ref{figure:hierarchy}.

\section{Coarse Category List.} \label{app:coarselist}
Table~\ref{tab:coarse} gives the details of the coarse-grained categories used during the annotation process, including the name, total number of fine-grained categories in each coarse-grained category, and some fine-grained examples for each coarse-grained category. 

\section{Category Descriptions Examples.} \label{app:desc}

In Table~\ref{tab:description}, we show examples of category descriptions in V3Det, which is written by human experts and chatgpt.

\begin{table*}[t]
	\begin{center}
            \tablestyle{12pt}{1.0}
\begin{tabular}{c|c|c|l}
\toprule
                     & Coarse category   &  Numbers            &  List of fine-grained category   \\ 
\midrule
\multirow{1}{*}{Coarse 1} & \multirow{1}{*}{Animal \& Human} & \multirow{1}{*}{7485} & siberian tiger; masai lion; net-winged insects; rugby player; man; woman; ...\\
\midrule
\multirow{1}{*}{Coarse 2} & \multirow{1}{*}{Device} & \multirow{1}{*}{241} & ceiling fan; tablet computer; display device; mobile device; ipad; game controller; ...\\
\midrule
\multirow{1}{*}{Coarse 3} & \multirow{1}{*}{Table \& Chair} & \multirow{1}{*}{30} & armchair; conference table; dining table; folding chair; gateleg table; ...\\
\midrule
\multirow{1}{*}{Coarse 4} & \multirow{1}{*}{Flower} & \multirow{1}{*}{1911} & hybrid tea rose; floribunda; tagetes; alpine aster; hawaiian hibiscus; orange lily; ...\\
\midrule
\multirow{2}{*}{Coarse 5} & Vegetables \& Beans & \multirow{2}{*}{696} & \multirow{2}{*}{squash; kiwifruit; saba banana; calamondin; matoke; eastern prickly pear; ...}\\
& \& Fruit \& Peel &&\\
\midrule
\multirow{3}{*}{Coarse 6} & Dishes \& Meat \& & \multirow{3}{*}{602} & \multirow{3}{*}{chicken meat; fried noodles; turkey meat; hot dog bun; sliced bread; soba; ...}\\
& Staple food \& Aggs &&\\
& Bean products \& Aggs &&\\
\midrule
\multirow{1}{*}{Coarse 7} & \multirow{1}{*}{Wearable Items} & \multirow{1}{*}{352} & baseball uniform; knit cap; martial arts uniform; dog clothes; diving mask; ...\\
\midrule
\multirow{1}{*}{Coarse 8} & \multirow{1}{*}{Fungus} & \multirow{1}{*}{301} & medicinal mushroom; lingzhi mushroom; pleurotus eryngii;russula fragilis; ...\\
\midrule
\multirow{1}{*}{Coarse 9} & \multirow{1}{*}{Vehicle} & \multirow{1}{*}{223} & microvan; general motors; steam car; solar vehicle; hoverboarding; ambulance; ...\\
\midrule
\multirow{2}{*}{Coarse 10} & \multirow{1}{*}{Sports equipment} & \multirow{2}{*}{168} & \multirow{2}{*}{skateboard truck; freebord; training bench; pommelhorse; dart; ...}\\
& w/o ball &&\\
\midrule
\multirow{3}{*}{Coarse 11} & \multirow{1}{*}{Drinks \& Seasoning \&} & \multirow{3}{*}{113} & \multirow{3}{*}{wiener melange; mocaccino; cream liqueur; ice cream; soy sauce; ...}\\
& Oil \& Dairy products \& &&\\
& Liquid chocolate &&\\
\midrule
\multirow{1}{*}{Coarse 12} & \multirow{1}{*}{Musical instrument} & \multirow{1}{*}{109} & drum stick; tabla; banjo; cymbal; drums; guitar; harp; piano; shekere; ...\\
\midrule
\multirow{1}{*}{Coarse 13} & \multirow{1}{*}{Hand \& Ignition tools} & \multirow{1}{*}{98} & hand fan; alligator wrench; bolt cutter; cap opener; corkscrew; forceps; ...\\
\midrule
\multirow{1}{*}{Coarse 14} & \multirow{1}{*}{Hardware gadgets} & \multirow{1}{*}{66} & keychain; plumbing fitting; threading needle; anchor; anvil; awl; bodkin; ...\\
\midrule
\multirow{2}{*}{Coarse 15} & Weapon w/o & \multirow{2}{*}{62} & \multirow{2}{*}{barbette carriage; battering ram; bomb; brass knucks; bullet; cannon; ...}\\
& hacking and cutting &&\\
\midrule
\multirow{2}{*}{Coarse 16} & Cards \& Paper products & \multirow{2}{*}{62} & \multirow{2}{*}{payment card; academic certificate; blackboard; bookmark; clipboard; doorplate; ...}\\
& \& Board &&\\
\midrule
\multirow{2}{*}{Coarse 17} & Cutting tool \& & \multirow{2}{*}{54} & \multirow{2}{*}{axe; bucksaw; crosscut saw; hedge trimmer; knife; cheese cutter; peeler; ...}\\
& Chop cold weapons &&\\
\midrule
\multirow{1}{*}{Coarse 18} & \multirow{1}{*}{Tableware w/o electricity} & \multirow{1}{*}{53} & wine glass; stemware; barbecue grill; food steamer; cup; chopstick; fork; ...\\
\midrule
\multirow{1}{*}{Coarse 19} & \multirow{1}{*}{Building \& Tent} & \multirow{1}{*}{51} & residential; water castle; portable toilet; cabana; bell tent; cenotaph; ...\\
\midrule
\multirow{1}{*}{Coarse 20} & \multirow{1}{*}{Personal care} & \multirow{1}{*}{50} & facial cleanser; baby powder; bottlebrush; condom; curler; hairbrush; ...\\
\midrule
\multirow{1}{*}{Coarse 21} & \multirow{1}{*}{Ball} & \multirow{1}{*}{48} & basketball; bowling ball; cricket ball; croquet ball; golf ball; handball; ...\\
\midrule
\multirow{1}{*}{Coarse 22} & \multirow{1}{*}{Cactus} & \multirow{1}{*}{39} & san pedro cactus; large-flowered cactus; ferocactus cylindraceus; ...\\
\midrule
\multirow{2}{*}{Coarse 23} & Window \& Door \& & \multirow{2}{*}{38} & \multirow{2}{*}{window covering; water well; artesian well; doorknob; dormer; gusher; ...}\\
& Delivery hole \& Well &&\\
\midrule
\multirow{1}{*}{Coarse 24} & \multirow{1}{*}{Cloth \& Cloth material} & \multirow{1}{*}{36} & folding napkins; bath mat; beach towel; cleaning pad; dustcloth; doily; ...\\
\midrule
\multirow{1}{*}{Coarse 25} & \multirow{1}{*}{Office equipment} & \multirow{1}{*}{37} & board eraser; abacus; chalk; fountain pen; marker; pencil; inkstone; ...\\
\midrule
\multirow{1}{*}{Coarse 26} & \multirow{1}{*}{Cleaning tools \& Nozzle} & \multirow{1}{*}{35} & bathroom sink; toilet roll holder; irrigation sprinkler; bathtub; broom; dumpster; ...\\
\midrule
\multirow{1}{*}{Coarse 27} & \multirow{1}{*}{Measuring equipment} & \multirow{1}{*}{28} & beaker; compass; detector; divider; plumb bob; protractor; triangular ruler; ...\\
\midrule
\multirow{1}{*}{Coarse 28} & \multirow{1}{*}{Shelf \& Storage cabinet} & \multirow{1}{*}{28} & wine rack; shoe organizer; pot rack; bookcase; clothes tree; coatrack; ...\\
\midrule
\multirow{1}{*}{Coarse 29} & \multirow{1}{*}{Gem \& Fountain} & \multirow{1}{*}{28} & fountain; crystal; diamond; ruby; pearl; emerald; chrysoberyl; jadeite; ...\\
\midrule
\multirow{1}{*}{Coarse 30} & \multirow{1}{*}{Faith related objects} & \multirow{1}{*}{27} & amulet; christian cross; flag; shoulder board; totem pole; medal; ...\\
\midrule
\multirow{1}{*}{Coarse 31} & \multirow{1}{*}{Medical related objects} & \multirow{1}{*}{26} & aspirator; catheter; hypodermic needle; pill; plaster; stethoscope; ...\\
\midrule
\multirow{1}{*}{Coarse 32} & \multirow{1}{*}{Pole \& Tube} & \multirow{1}{*}{24} & blowgun; chimney; cigar; fire hose; grab bar; meerschaum; test tube; ...\\
\midrule
\multirow{1}{*}{Coarse 33} & \multirow{1}{*}{Lifesaving objects} & \multirow{1}{*}{21} & baby float; air cushion; breeches buoy; fire hydrant; life buoy; water wings; ...\\
\midrule
\multirow{2}{*}{Coarse 34} & Bottles \& Bags \& & \multirow{2}{*}{22} & \multirow{2}{*}{weaving basket; bag; barrel; basin; bottle; briefcase; pot; rain barrel; ...}\\
& Buckets \& Boxes &&\\
\midrule
\multirow{1}{*}{Coarse 35} & \multirow{1}{*}{Candy \& Solid chocolate} & \multirow{1}{*}{21} & brittle; chewing gum; candied apple; cocoa powder; lollipop; jello; ...\\
\midrule
\multirow{1}{*}{Coarse 36} & \multirow{1}{*}{Planet \& Satellite} & \multirow{1}{*}{18} & satellite; meteorite; moon; sun; black hole; Earth; Mars; Mercury; Venus; ...\\
\midrule
\end{tabular}
\end{center}
\end{table*}

\begin{table*}[t]
	\begin{center}
            \tablestyle{12pt}{1.0}
\begin{tabular}{c|c|c|l}
\toprule
                     & Coarse category   &  Numbers            &  List of fine-grained categories   \\ 
\midrule
\multirow{1}{*}{Coarse 37} & \multirow{1}{*}{Bedding} & \multirow{1}{*}{18} & infant bed; bed; bed pillow; bunk bed; carrycot; crib; futon; hammock; ...\\
\midrule
\multirow{1}{*}{Coarse 38} & \multirow{1}{*}{Chess} & \multirow{1}{*}{16} & chessman; chess set; Chinese Chess; Go; International Draughts or Checkers; Shogi; ...\\
\midrule
\multirow{1}{*}{Coarse 39} & \multirow{1}{*}{Bell} & \multirow{1}{*}{12} & electric bell; timer; weathervane; wind chime; fire alarm; Windbell; bicycle bell; ...\\
\midrule
\multirow{1}{*}{Coarse 40} & \multirow{1}{*}{Signpost \& Roadblock} & \multirow{1}{*}{12} & stop sign; crosswalk sign; billboard; pedestrian crossing; yard marker; ...\\
\midrule
\multirow{3}{*}{Coarse 41} & Control device \& Heating & \multirow{3}{*}{12} & \multirow{3}{*}{brazier; gearshift; handwheel; hot-water bottle; radiator; roaster; ...}\\
&  appliances \& Hot-water &&\\
& bag w/o electricity &&\\
\midrule
\multirow{1}{*}{Coarse 42} & \multirow{1}{*}{Wheel shaped objects} & \multirow{1}{*}{11} & bicycle wheel; bobbin; ferris wheel; gear; inner tube; pulley; automotive tire; ...\\
\midrule
\multirow{1}{*}{Coarse 43} & \multirow{1}{*}{Lens} & \multirow{1}{*}{10} & magnifying glass; microscope; telescope; Rifle scopes; Triangular Prism; ...\\
\midrule
\multirow{1}{*}{Coarse 44} & \multirow{1}{*}{Currency \& Whistle} & \multirow{1}{*}{8} & whistle; gold; money; cash; paper money; coinage; banknote; ...\\
\midrule
\multirow{1}{*}{Coarse 45} & \multirow{1}{*}{Animal nest} & \multirow{1}{*}{9} & birdcage; chicken coop; rabbit hutch; nest; wasp nest; cage; ...\\
\midrule
\multirow{1}{*}{Coarse 46} & \multirow{1}{*}{Umbrella \& Ladder} & \multirow{1}{*}{9} & parachute; cocktail umbrella; Aluminum alloy ladder; Wooden ladder; ...\\
\midrule
\multirow{1}{*}{Coarse 47} & \multirow{1}{*}{Electronic component} & \multirow{1}{*}{6} & battery; capacitor; coil; resistor; solar cell; electronic component; ...\\
\midrule
\multirow{1}{*}{Coarse 48} & \multirow{1}{*}{Socket \& Plug} & \multirow{1}{*}{6} & bung; cork; power outlet; socket; wall socket; ...\\
\midrule
\multirow{2}{*}{Coarse 49} & Lure \& Aquarium & \multirow{2}{*}{6} & \multirow{2}{*}{aquarium; fishbowl; pillar box; Nano aquarium; Spoon lure; Penfold post box; ...}\\
& \& Post box &&\\
\midrule
\multirow{2}{*}{Coarse 50} &Industrial machine & \multirow{2}{*}{5} & \multirow{2}{*}{concrete mixer; crane; generator; spray paint; pumpjack; ...}\\
& \& Spray paint &&\\
\midrule
\multirow{2}{*}{Coarse 51} & Spring \& Magnet & \multirow{2}{*}{4} & \multirow{2}{*}{compass; rubber band; spring; refrigerator magnet; ...}\\
& \& Compass &&\\
\midrule
\multirow{3}{*}{Coarse 52} & Popcorn machine & \multirow{3}{*}{5} & \multirow{3}{*}{ashtray; automated teller machine; censer; popper; incense burner; ...}\\
& ATM \& Ashtray \& &&\\
& Incense burner &&\\
\midrule
\multirow{1}{*}{Coarse 53} & \multirow{1}{*}{Model} & \multirow{1}{*}{93} & toy vehicle; rubik's cube; amphora; armillary sphere; cockhorse; doll; ...\\
\bottomrule
\end{tabular}
	\end{center}
    \vspace{-10pt}
 	\caption{
		\small{\textbf{The details of the coarse-grained categories.} All fine-grained categories are divided into 53 coarse-grained categories. `Numbers' denotes total number of fine-grained categories in each coarse-grained category.
		}
	}\label{tab:coarse}
\end{table*}

\begin{table*}[t]
	\begin{center}
            \tablestyle{10pt}{1.5}
            \begin{tabular}{c|c|c|l}
            \toprule
             Name & Image & Type & Description \\
            \midrule
            \multirow{2}{*}{\makecell{platycercus \\ adscitus}} & \multirow{2}{*}{\includegraphics[height=1cm]{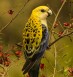}} & Experts & platycercus adscitus has a brown head, yellow belly and green wings \\
            & & chatgpt & green body, yellow head, red markings (pale-headed rosella) \\
            \midrule
            \multirow{2}{*}{\makecell{Adelie \\ penguin}} & \multirow{2}{*}{\includegraphics[height=1cm]{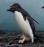}} & Experts & medium-sized penguins occurring in large colonies on the Adelie Coast of Antarctica \\
            & & chatgpt & the Adelie penguin has a black head and back, white front, and a distinctive white ring around the eye \\
            \midrule
            \multirow{2}{*}{polar bear} & \multirow{2}{*}{\includegraphics[height=1cm]{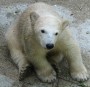}} & Experts & compared to other bears, smaller head, round ears, slender neck. Black skin, huge and ferocious. \\
            & & chatgpt & large, white-furred bear, with a long neck, a stocky body, powerful legs, and sharp claws \\
            \midrule
            \multirow{2}{*}{pembroke} & \multirow{2}{*}{\includegraphics[height=1cm]{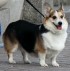}} & Experts & the smaller and straight-legged variety of corgi having pointed ears and a short tail \\
            & & chatgpt & small, sturdy dog breed, with erect ears, a foxy face, and a docked tail (or naturally bobtail) \\ 
            \midrule
            \multirow{2}{*}{\makecell{bengal \\ tiger}} & \multirow{2}{*}{\includegraphics[height=1cm]{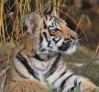}} & Experts & yellow-orange coats with dark stripes, white belly and limbs, and an orange tail with black rings \\
            & & chatgpt & orange-brown coat with black stripes, white belly, and distinctive white markings above the eyes \\ 
            \midrule
            \multirow{2}{*}{polar hare} & \multirow{2}{*}{\includegraphics[height=1cm]{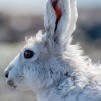}} & Experts & a large hare of northern North America; it is almost completely white in winter \\
            & & chatgpt & large, white-furred hare with long ears and strong hind legs, found in Arctic regions \\ 
            \midrule
            \multirow{2}{*}{gorilla} & \multirow{2}{*}{\includegraphics[height=1cm]{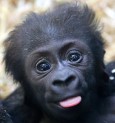}} & Experts & the gorilla is strong and hairless on its face and ears. it has a high forehead and protruding jawbone \\
            & & chatgpt & large, muscular primate with black or dark brown fur, a broad chest, and a prominent brow ridge \\ 
            \midrule
\end{tabular}
\end{center}
\end{table*}

\begin{table*}[t]
	\begin{center}
            \tablestyle{2pt}{1.5}
\begin{tabular}{c|c|c|l}
            \toprule
             Name & Image & Type & Description \\
            \midrule
            \multirow{2}{*}{\makecell{African \\ elephant}} & \multirow{2}{*}{\includegraphics[height=1cm]{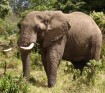}} & Experts & an elephant native to Africa having enormous flapping ears and ivory tusks \\
            & & chatgpt & massive, grey mammal with a long trunk, large ears, curved tusks, and wrinkled skin \\ 
            \midrule
            \multirow{2}{*}{charolais} & \multirow{2}{*}{\includegraphics[height=1cm]{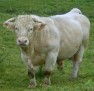}} & Experts & this breed of cattle is heavy, with weighing 700-1650 kg. they have white/cream coats and pink noses \\
            & & chatgpt & large, white-colored breed of cattle with a muscular build, broad forehead, and short horns \\ 
            \midrule
            \multirow{2}{*}{\makecell{Grevy's \\ zebra}} & \multirow{2}{*}{\includegraphics[height=1cm]{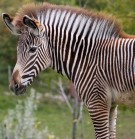}} & Experts & it resembles a mule with a big head, elongated nostril openings, large round conical ears, and tall erect mane \\
            & & chatgpt & large, wild equid with black and white stripes, a white belly, and a tall, erect mane \\ 
            \midrule
            \multirow{2}{*}{box turtle} & \multirow{2}{*}{\includegraphics[height=1cm]{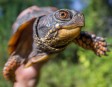}} & Experts & it has a domed shell hinged at the bottom, allowing them to close it tightly and escape predators \\
            & & chatgpt & small, domed shell with brown or olive-colored skin and a pattern of yellow or orange spots or lines \\ 
            \midrule
            \multirow{2}{*}{tuatara} & \multirow{2}{*}{\includegraphics[height=1cm]{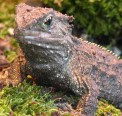}} & Experts & a reptile with a greenish-grey lizard-like appearance found only on certain small islands near New Zealand \\
            & & chatgpt & a reptile with a spiny crest, two rows of teeth, and a third "eye" on the top of its head \\ 
            \midrule
            \multirow{2}{*}{\makecell{northern \\ seahorse}} & \multirow{2}{*}{\includegraphics[height=1cm]{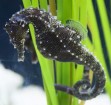}} & Experts & northern seahorses have long snouts and a curly tail, and can grow up to 8 inches in length \\
            & & chatgpt & the northern seahorse has a small body covered in bony plates, a long snout, and a curled tail \\ 
            \midrule
            \multirow{2}{*}{\makecell{lowland \\ burrowing \\ treefrog}} & \multirow{2}{*}{\includegraphics[height=1cm]{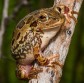}} & Experts & terrestrial burrowing nocturnal frog of grassy terrain and scrub forests having very hard upper surface of head \\
            & & chatgpt & small, round-bodied frog with smooth skin, adapted for burrowing with short legs and a pointed snout \\ 
            \midrule
            \multirow{2}{*}{rock beauty} & \multirow{2}{*}{\includegraphics[height=1cm]{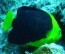}} & Experts & 
            predominately black. The head and front half portion of the body, and the caudal fin are a bright yellow \\
            & & chatgpt & 
            bright yellow head, tail, pectoral fins; black body, dorsal, anal fins; abrupt color transitions \\
            \midrule
            \multirow{2}{*}{clingfish} & \multirow{2}{*}{\includegraphics[height=1cm]{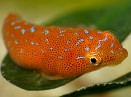}} & Experts & very small (to 3 inches) flattened marine fish with a sucking disc on the abdomen for clinging to rocks etc \\
            & & chatgpt & the clingfish has a flattened body with a suction cup-like pelvic disc, and ranges in color from brown to green \\ 
            \midrule
            \multirow{2}{*}{porpoise} & \multirow{2}{*}{\includegraphics[height=1cm]{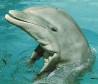}} & Experts & 
            whales distinguishable from dolphins by their more compact build, smaller size, and curved blunt snout with spatulate teeth \\
            & & chatgpt & small, grey marine mammal with a rounded head, a small dorsal fin, and a curved mouth \\ 
            \midrule
            \multirow{2}{*}{manta} & \multirow{2}{*}{\includegraphics[height=1cm]{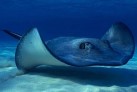}} & Experts & extremely large pelagic tropical ray with triangular pectoral fins, horn-shaped cephalic fins and large, forward-facing mouths \\
            & & chatgpt & large, flat-bodied, wing-like fins, no tail, black or dark upper side, white or light underside \\ 
            \midrule
            \multirow{2}{*}{\makecell{aglais}} & \multirow{2}{*}{\includegraphics[height=1cm]{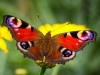}} & Experts & the wings of aglais are rusty red with a unique eyespot in black, blue, and yellow at each wingtip \\
            & & chatgpt & eyespots on wings, blue-green-brown hues (European peacock butterfly) \\ 
            \midrule
            \multirow{2}{*}{\makecell{woodland \\ sunflower}} & \multirow{2}{*}{\includegraphics[height=1cm]{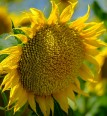}} & Experts & a wildflower native to the United States, which has bright yellow petals that surround a dark brown center \\
            & & chatgpt & tall yellow flower with numerous thin petals surrounding a dark center disk, and green leaves \\ 
            \midrule
            \multirow{2}{*}{calliandra} & \multirow{2}{*}{\includegraphics[height=1cm]{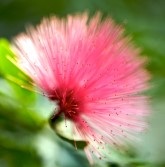}} & Experts & the flowers are produced in cylindrical or globose inflorescences and have numerous long slender stamens \\
            & & chatgpt & small shrub with fern-like leaves and vibrant, pink, powder-puff shaped flowers \\ 
            \midrule
            \multirow{2}{*}{\makecell{cistus \\ salviifolius}} & \multirow{2}{*}{\includegraphics[height=1cm]{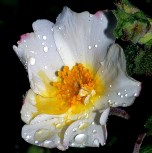}} & Experts & a shrub with fragrant, silver-green foliage and white flowers \\
            & & chatgpt & a tall, bright yellow or orange flower often grown for its edible seeds and as an ornamental plant \\ 
            \midrule
            \multirow{2}{*}{\makecell{power drill}} & \multirow{2}{*}{\includegraphics[height=1cm]{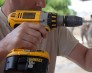}} & Experts & 
            a hand tool with a rotating chuck driven by an electric motor for drilling \\
            & & chatgpt & handheld tool with a motor and rotating chuck for drilling holes and fastening screws, often with a pistol grip \\ 
            \midrule
            \multirow{2}{*}{clarinet} & \multirow{2}{*}{\includegraphics[height=1cm]{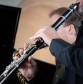}} & Experts & 
            the clarinet is a single-reed instrument with a nearly cylindrical bore and a flared bell. \\
            & & chatgpt & narrow, cylindrical woodwind instrument with a mouthpiece, reed, and numerous keys for playing different notes \\ 
            \midrule
            \multirow{2}{*}{\makecell{off-road \\ vehicle}} & \multirow{2}{*}{\includegraphics[height=1cm]{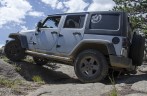}} & Experts & the off-road vehicles typically have four-wheel-drive, increased suspension, and large tires \\
            & & chatgpt & large, sturdy vehicle with high ground clearance, wide tires, and typically four-wheel drive for use on unpaved terrain \\ 
            \bottomrule
            \end{tabular}
	\end{center}
        \vspace{-15pt}
 	\caption{
		Examples of category descriptions of V3Det.
	}\label{tab:description}
	\vspace{-10pt}
\end{table*}

\section{More Dataset Visualizations.} \label{app:vis}
\begin{figure*}[t]
    \center
    \small
    \setlength\tabcolsep{1pt}
    {
    \begin{tabular}{cc}
        \includegraphics[width=1\columnwidth, height=5.5cm]{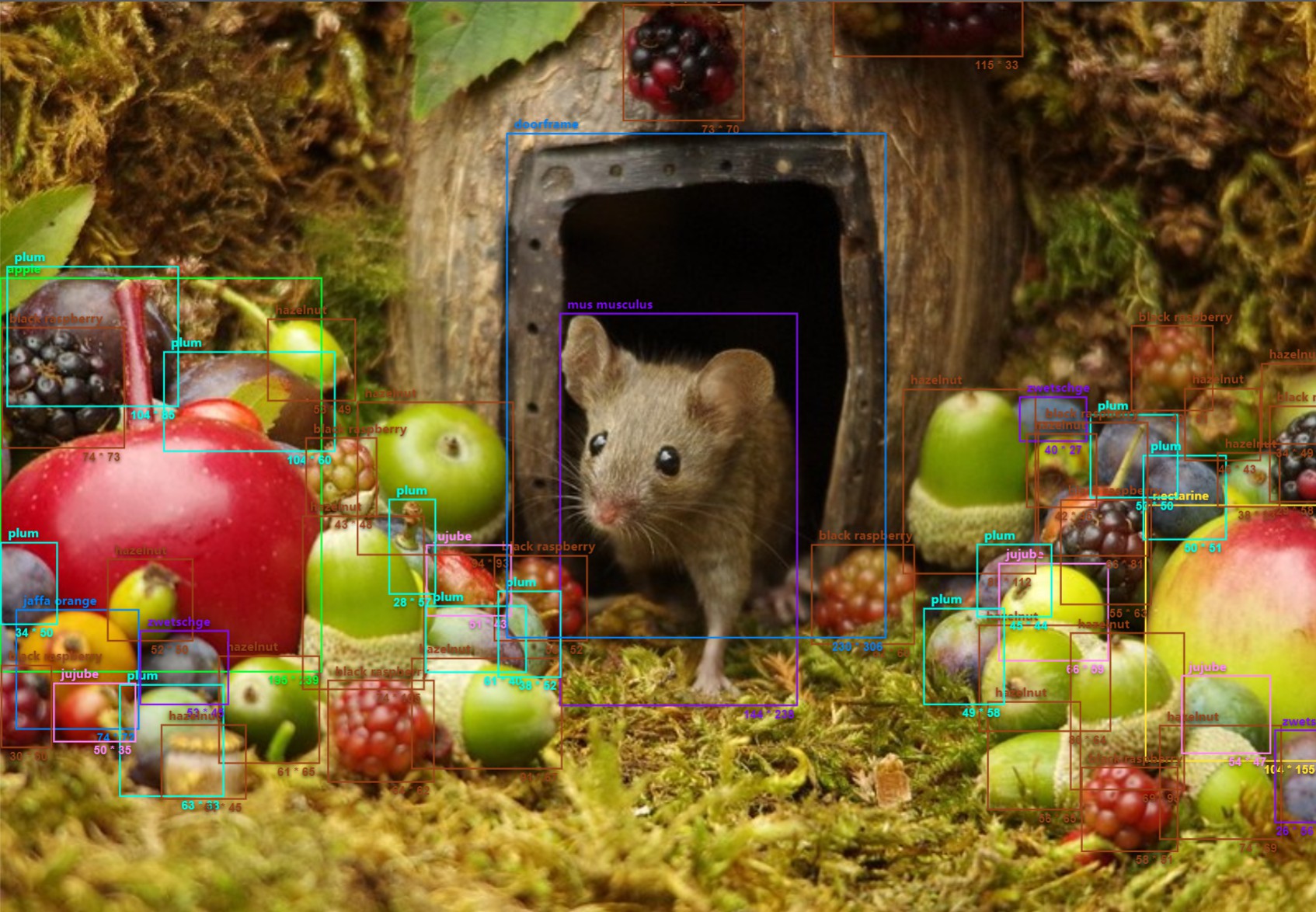} &
        \includegraphics[width=1\columnwidth, height=5.5cm]{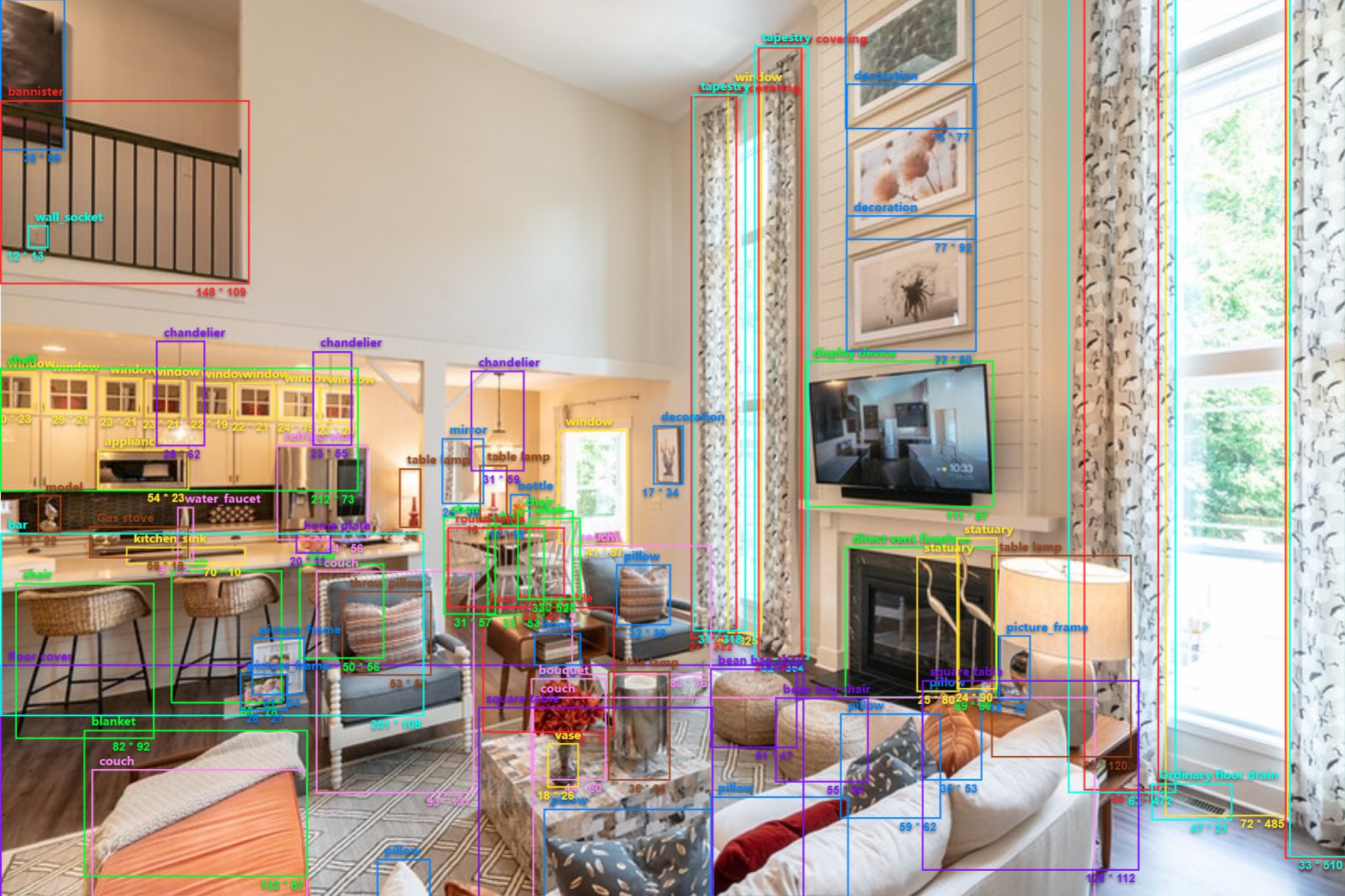} \\
        \includegraphics[width=1\columnwidth, height=5.5cm]{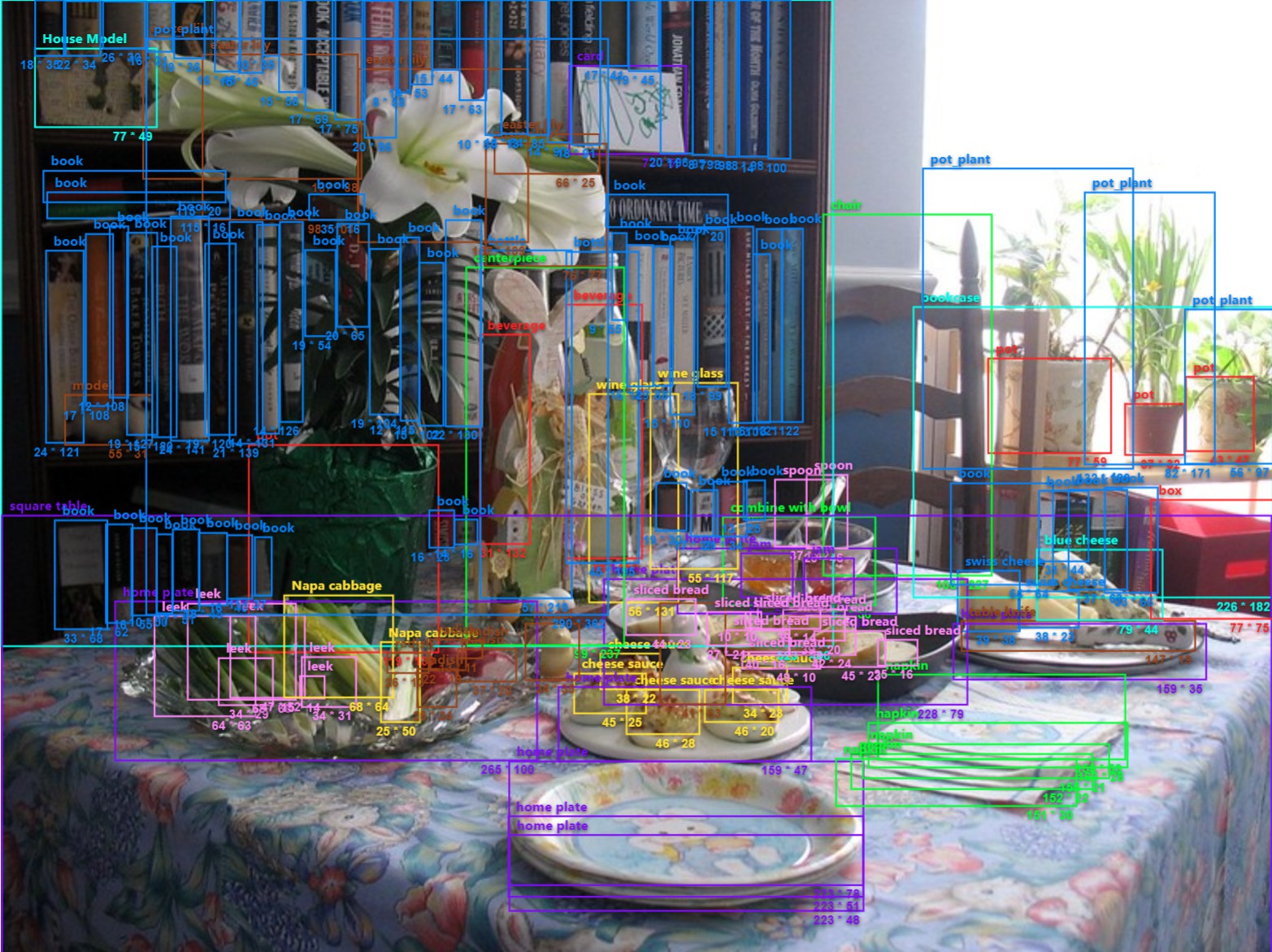} &
        \includegraphics[width=1\columnwidth, height=5.5cm]{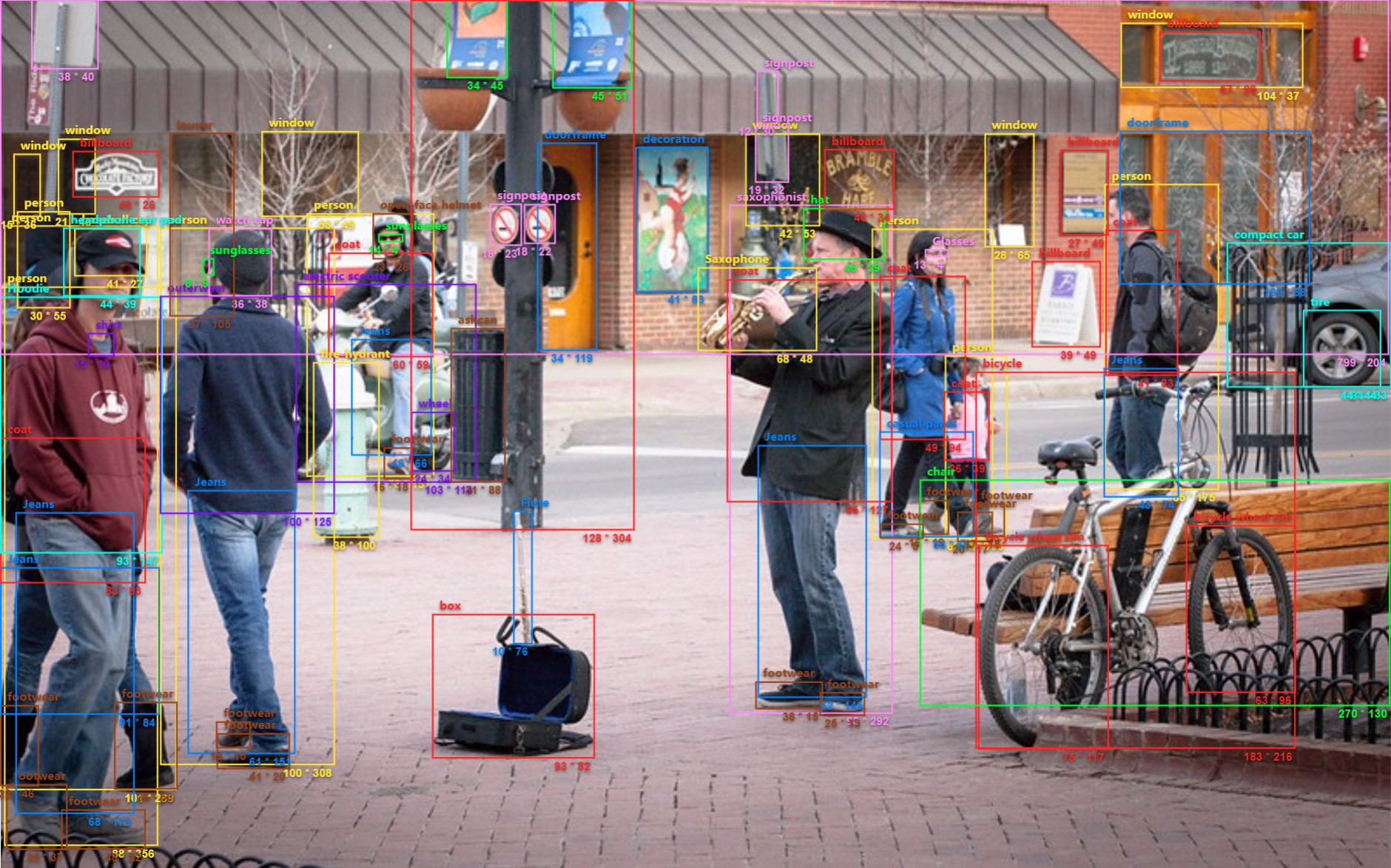} \\
        \includegraphics[width=1\columnwidth, height=5.5cm]{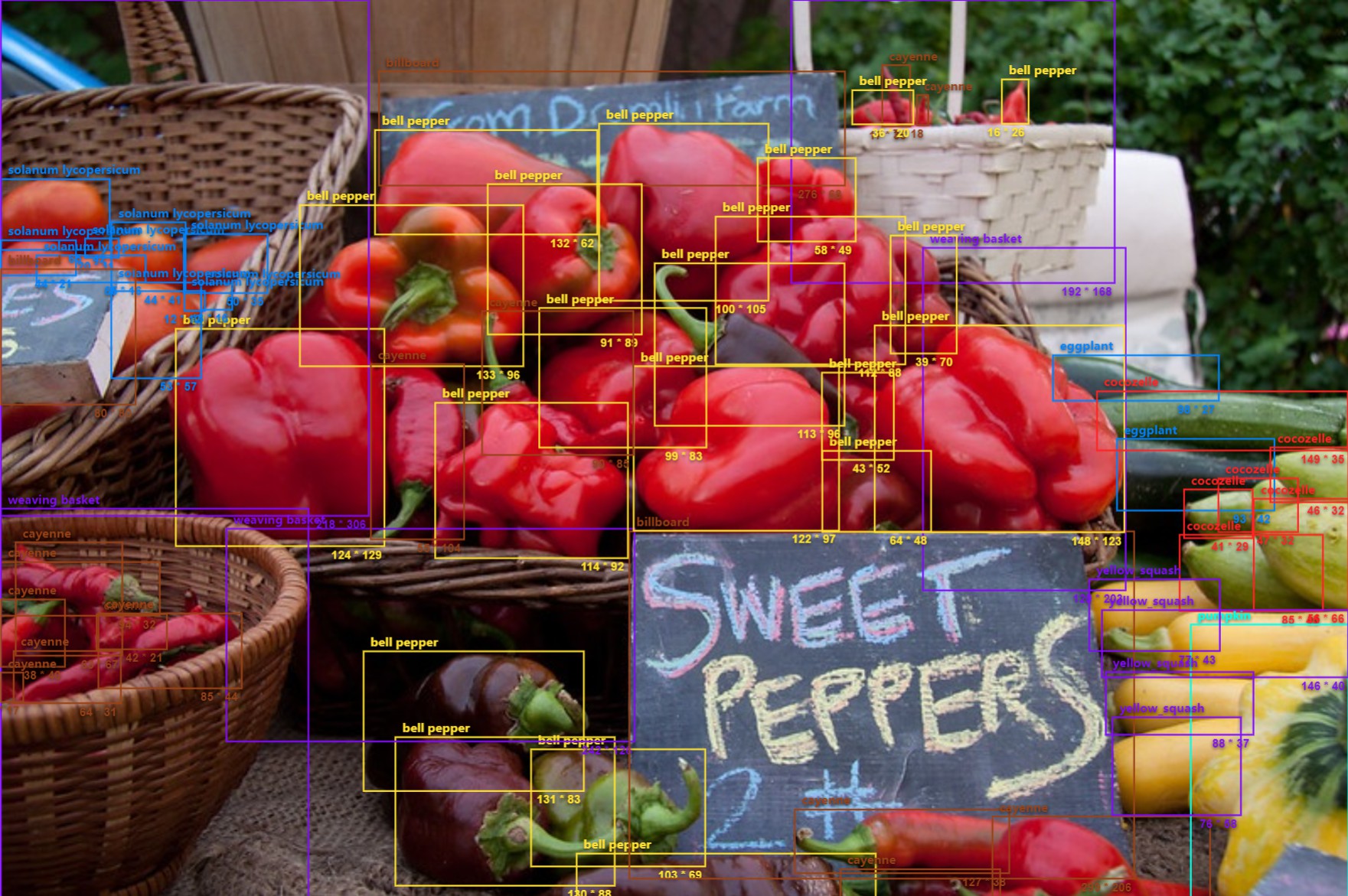} &
        \includegraphics[width=1\columnwidth, height=5.5cm]{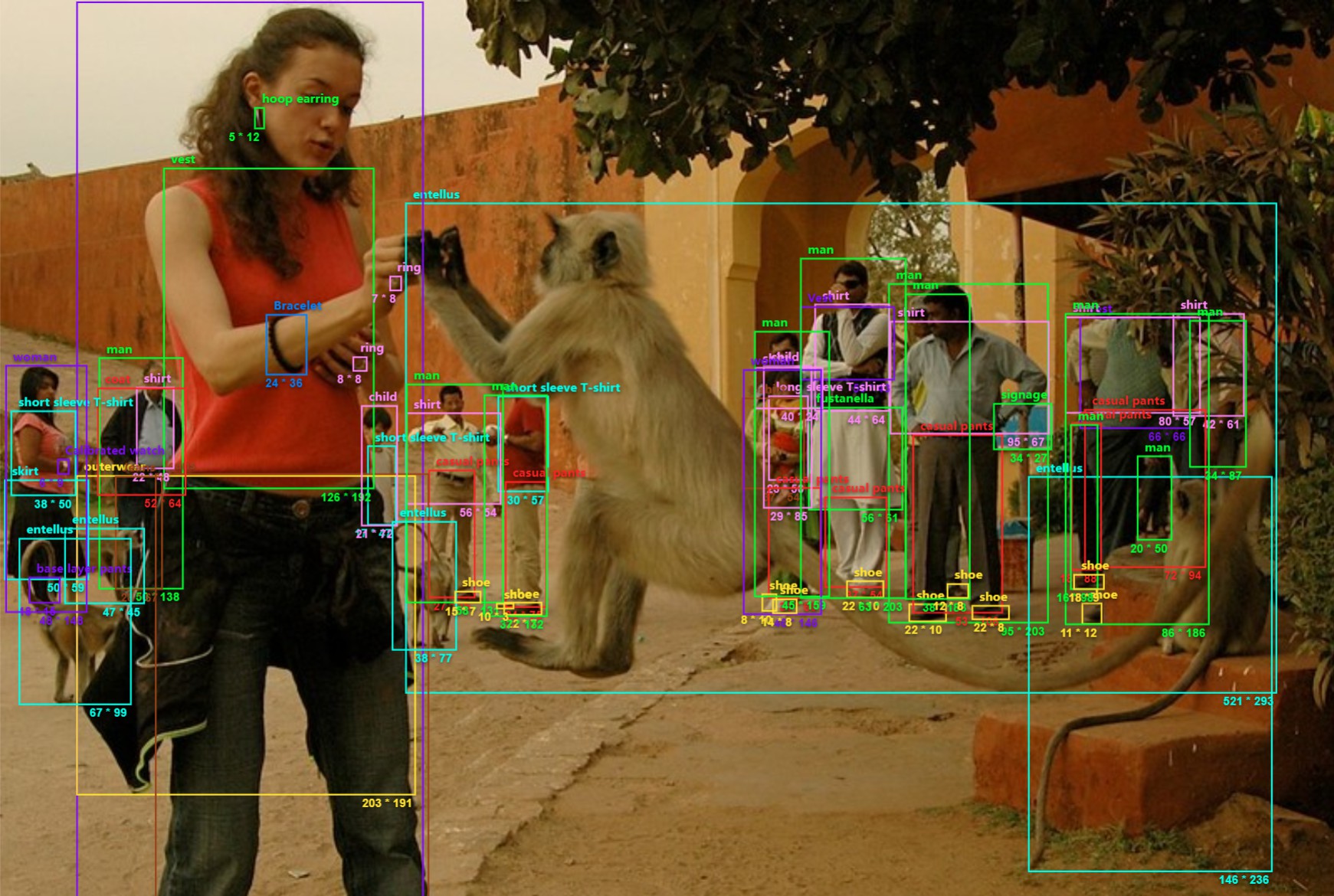} \\
        \includegraphics[width=1\columnwidth, height=5.5cm]{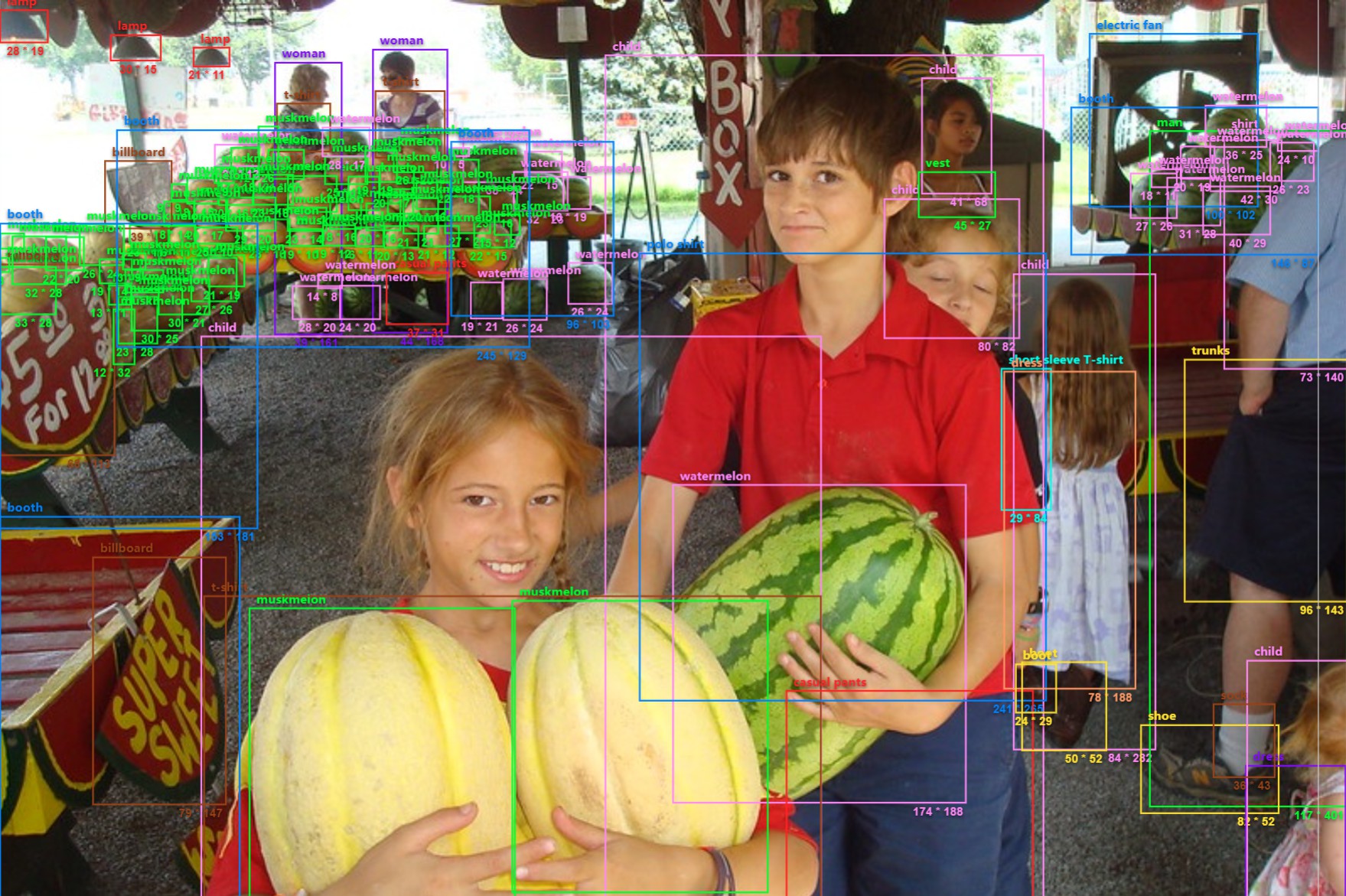} &
        \includegraphics[width=1\columnwidth, height=5.5cm]{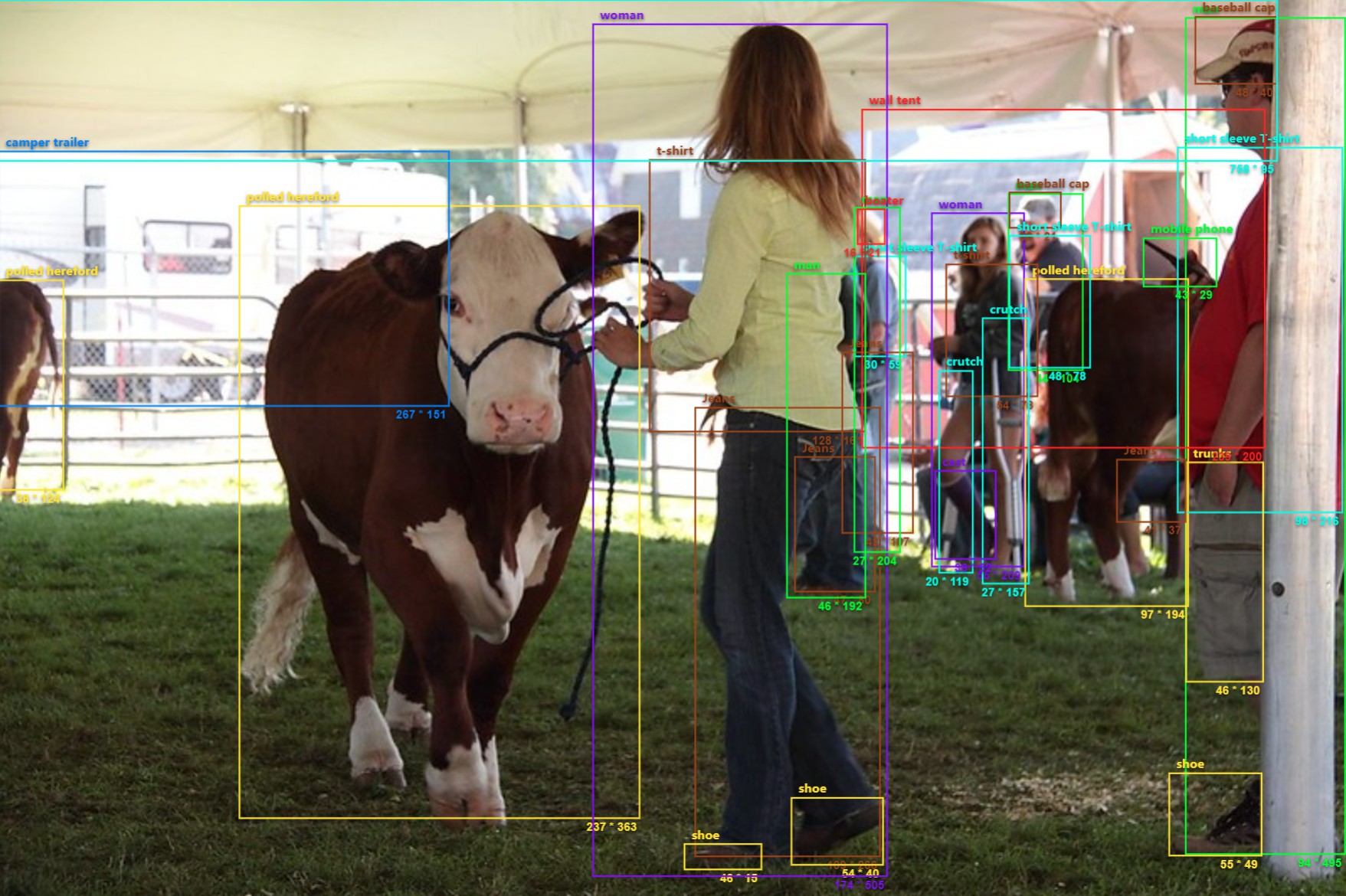} \\
    \end{tabular}
    }
    \vspace{-5pt}
\caption{\textbf{Visualizations of annotations in complex scenes.} Each box is paired with category name and box size.}
\label{figure:vis_complex}
\vspace{-7pt}
\end{figure*}

\begin{figure*}[t]
    \center
    \small
    \setlength\tabcolsep{1pt}
    {
    \begin{tabular}{cccc}
        \includegraphics[height=3.2cm]{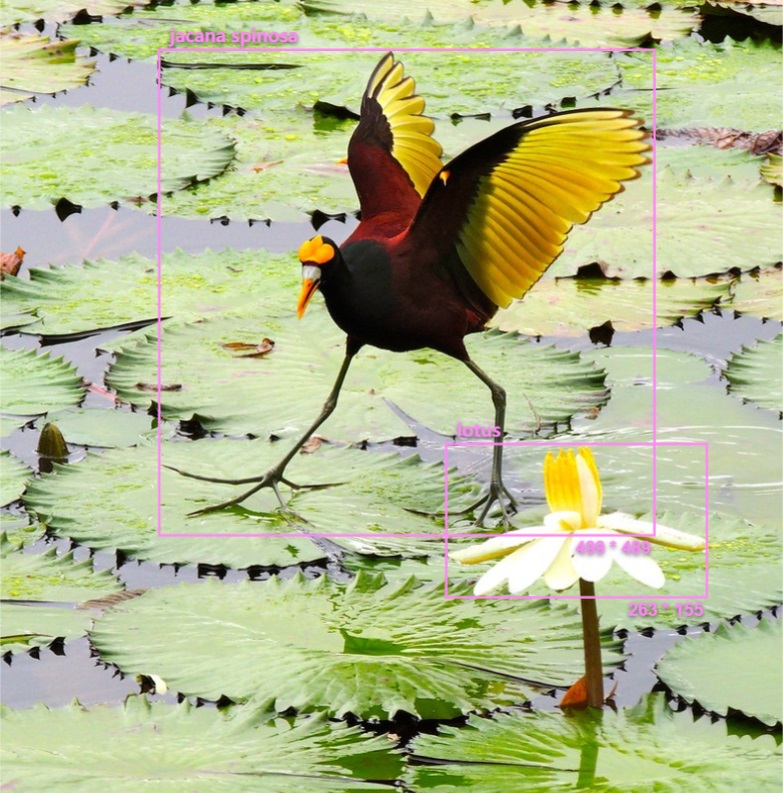} &
        \includegraphics[height=3.2cm]{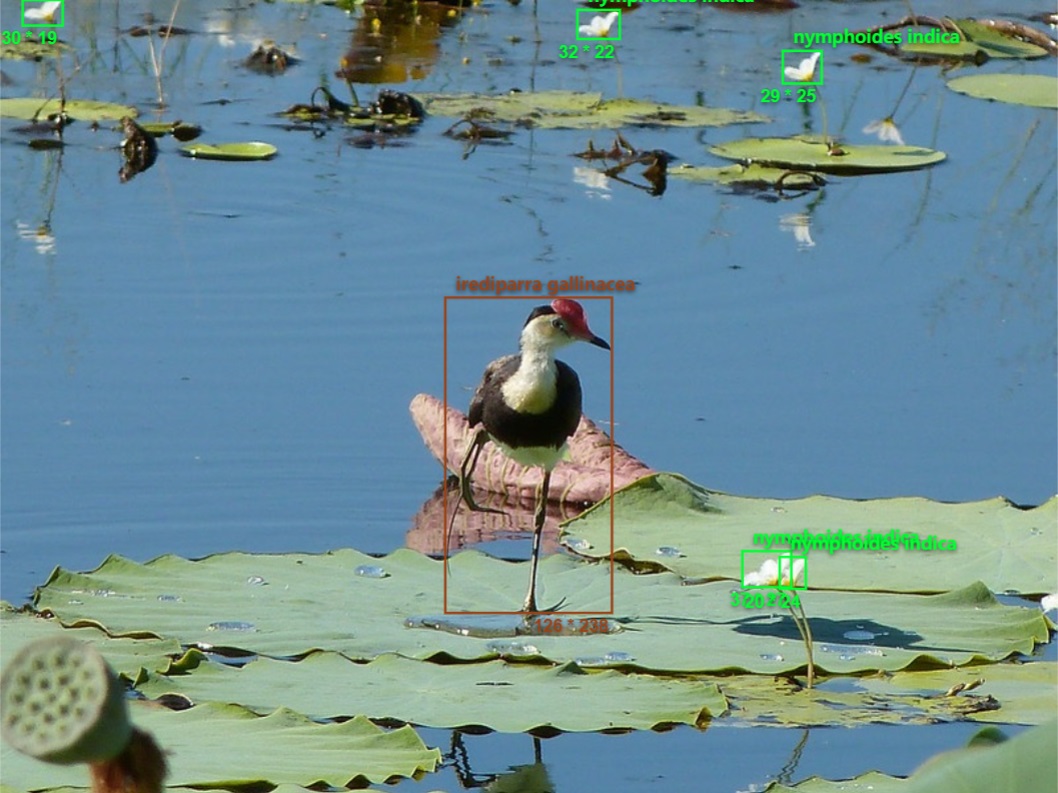} &
        \includegraphics[height=3.2cm]{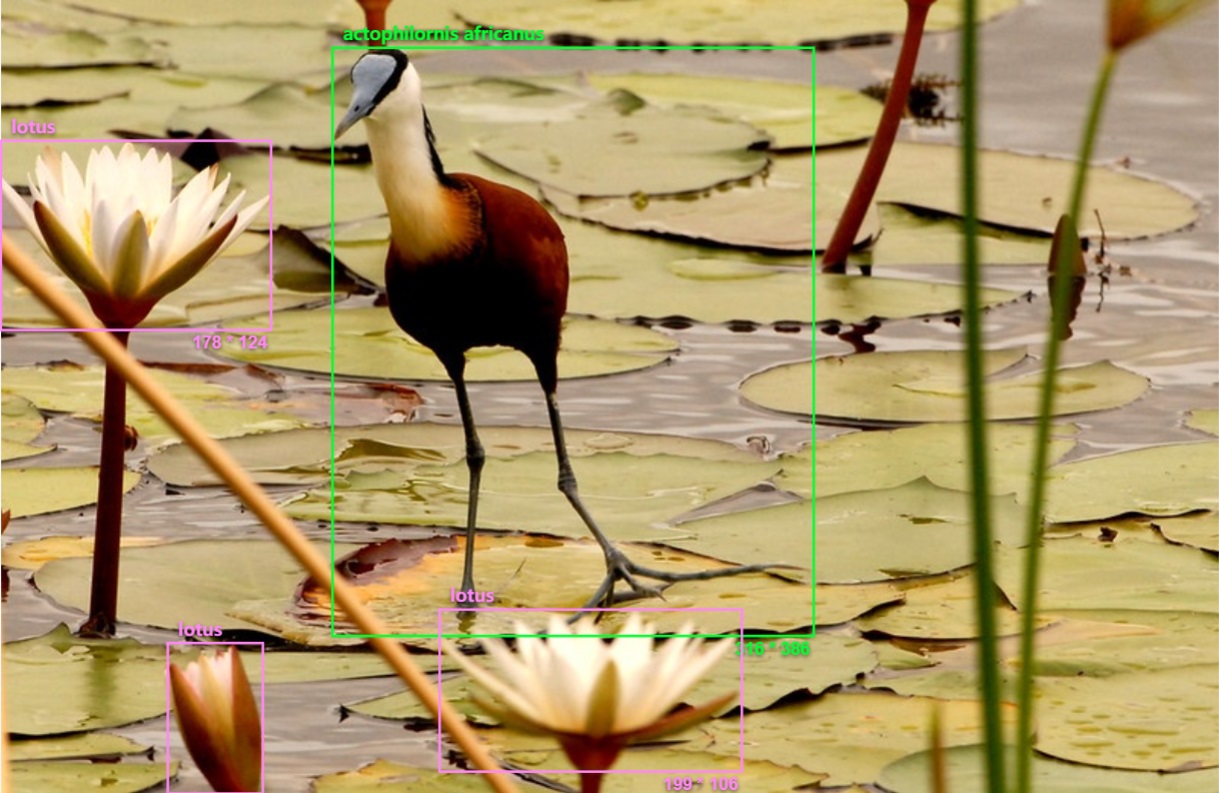} &
        \includegraphics[height=3.2cm]{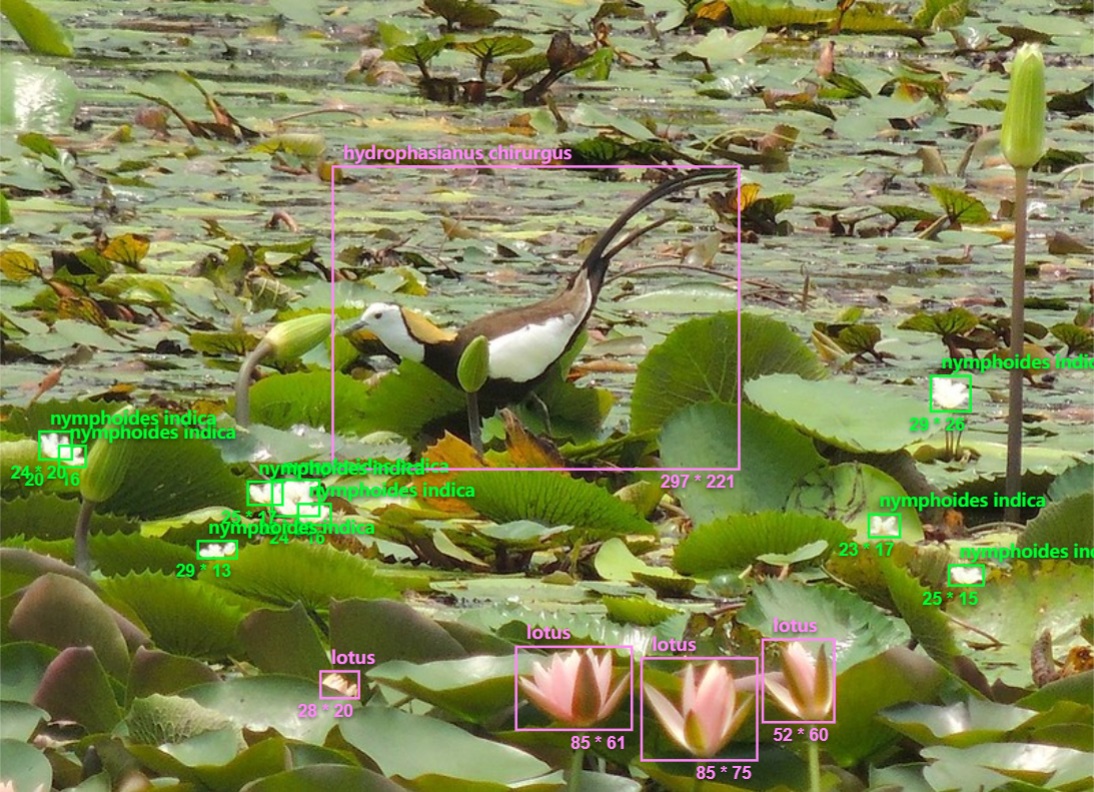} \\
    \end{tabular} \\
    \begin{tabular}{cccc}
        \includegraphics[height=3.26cm]{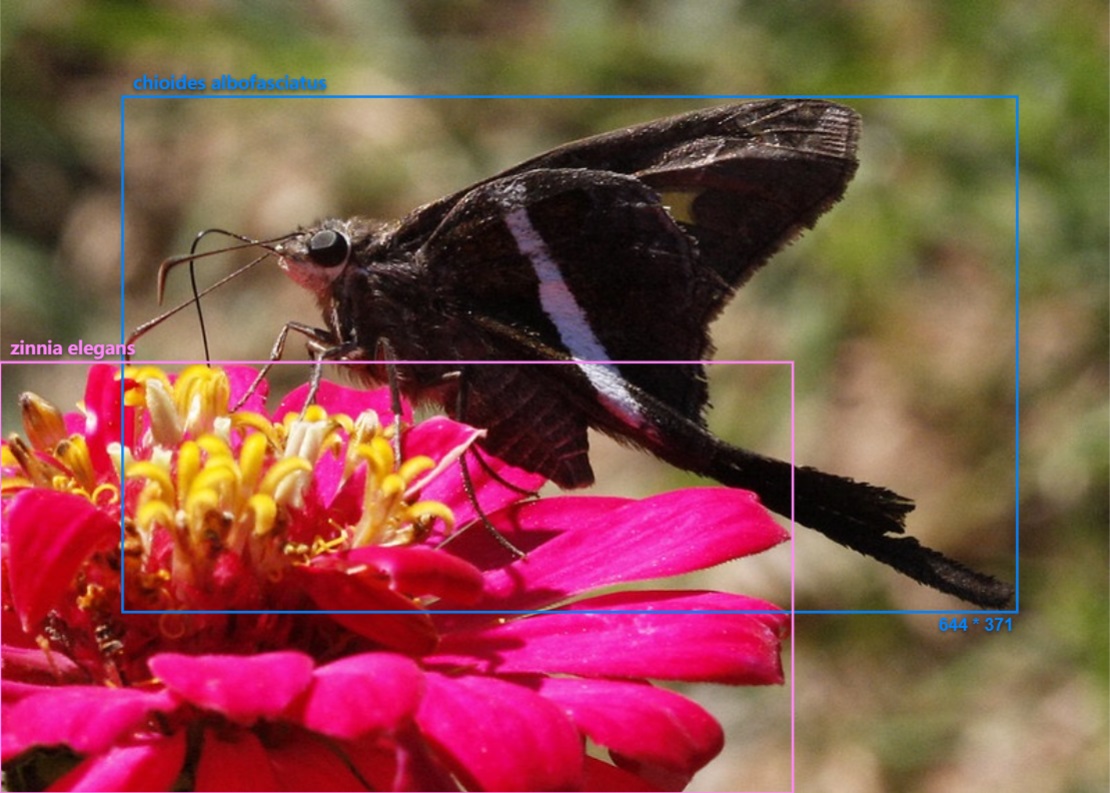} &
        \includegraphics[height=3.26cm]{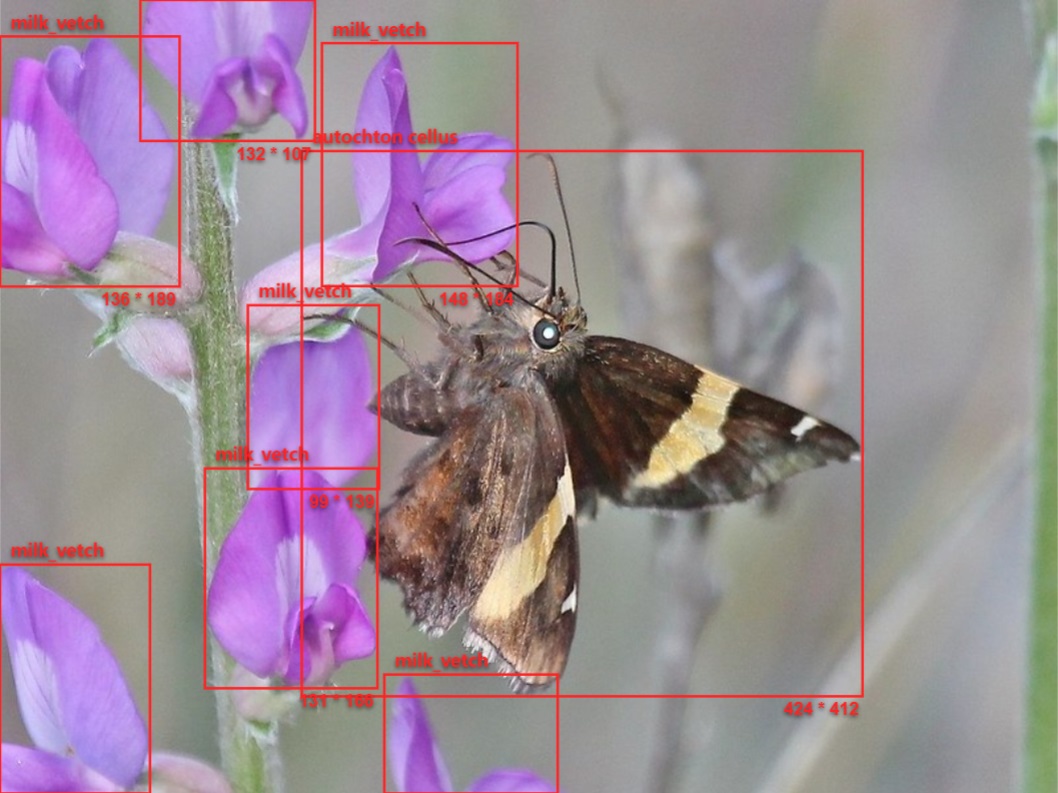} &
        \includegraphics[height=3.26cm]{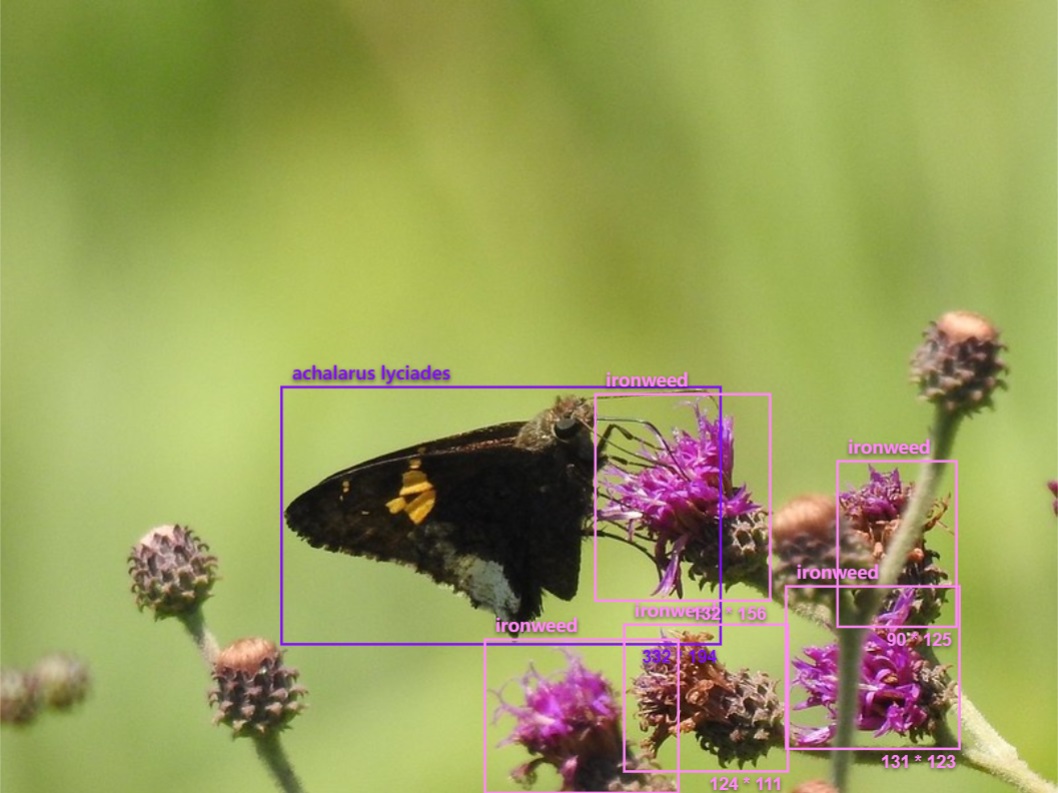} &
        \includegraphics[height=3.26cm]{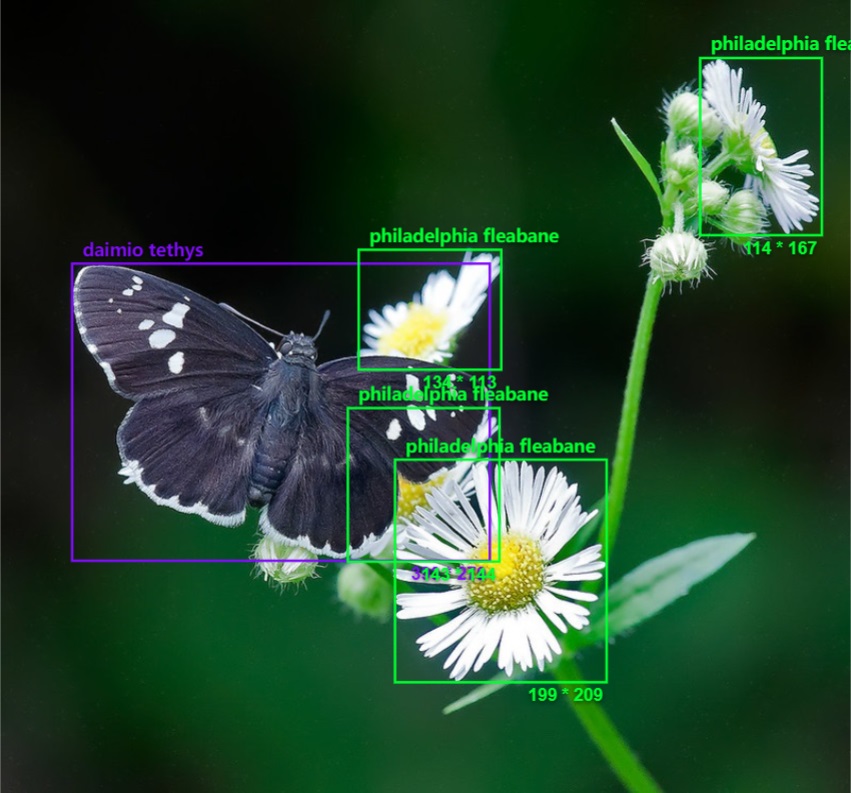} \\
    \end{tabular} \\
    \begin{tabular}{cccc}
        \includegraphics[height=3.05cm]{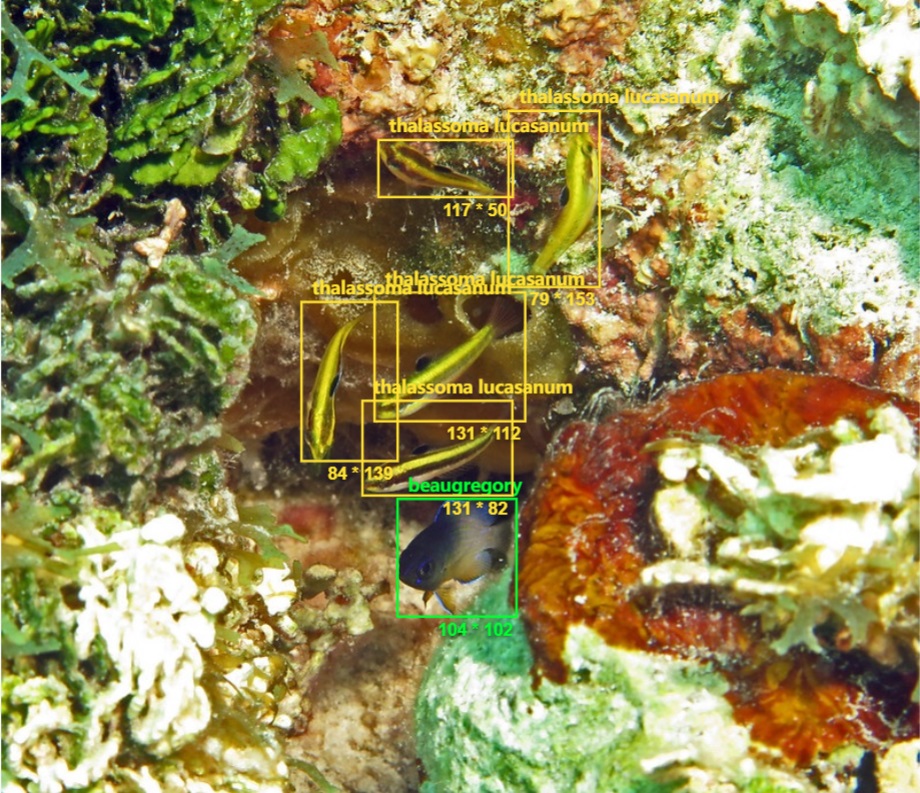} &
        \includegraphics[height=3.05cm]{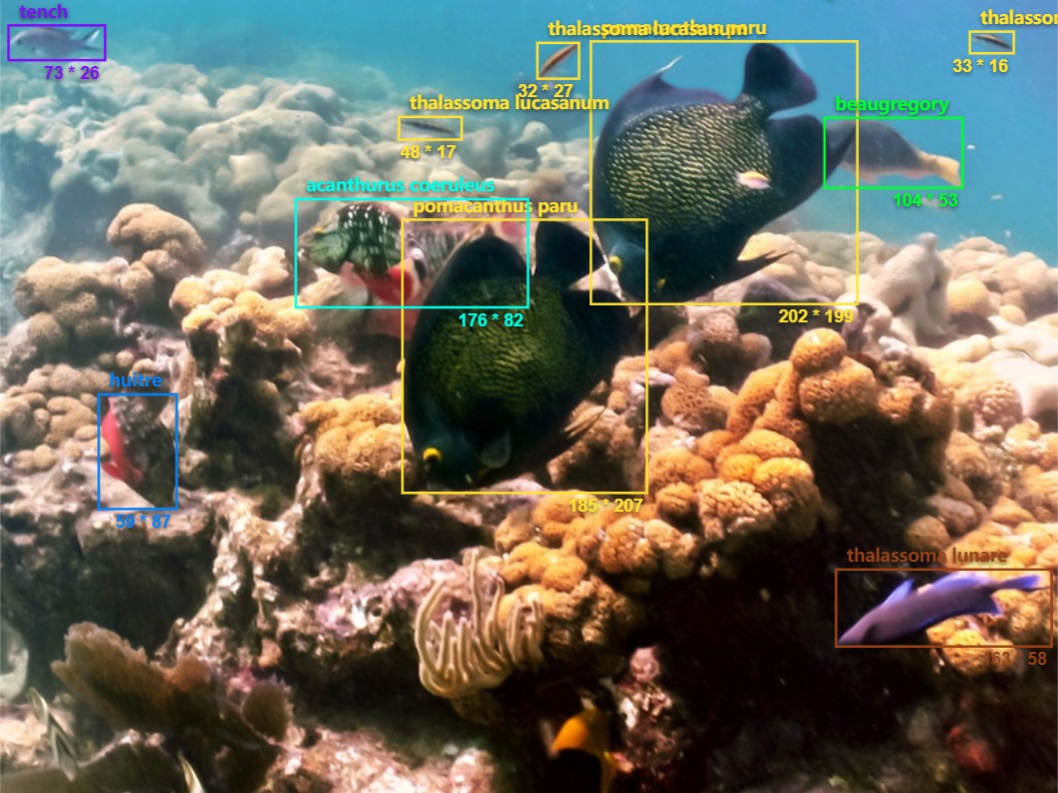} &
        \includegraphics[height=3.05cm]{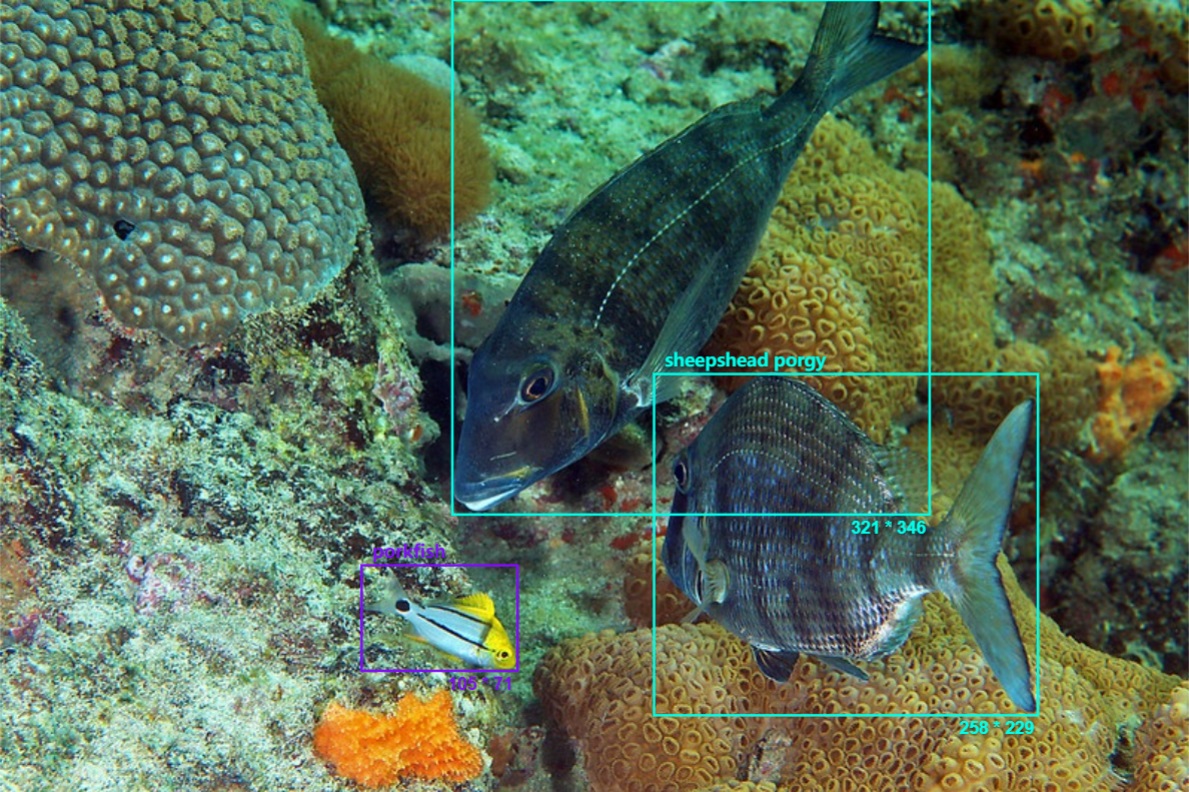} &
        \includegraphics[height=3.05cm]{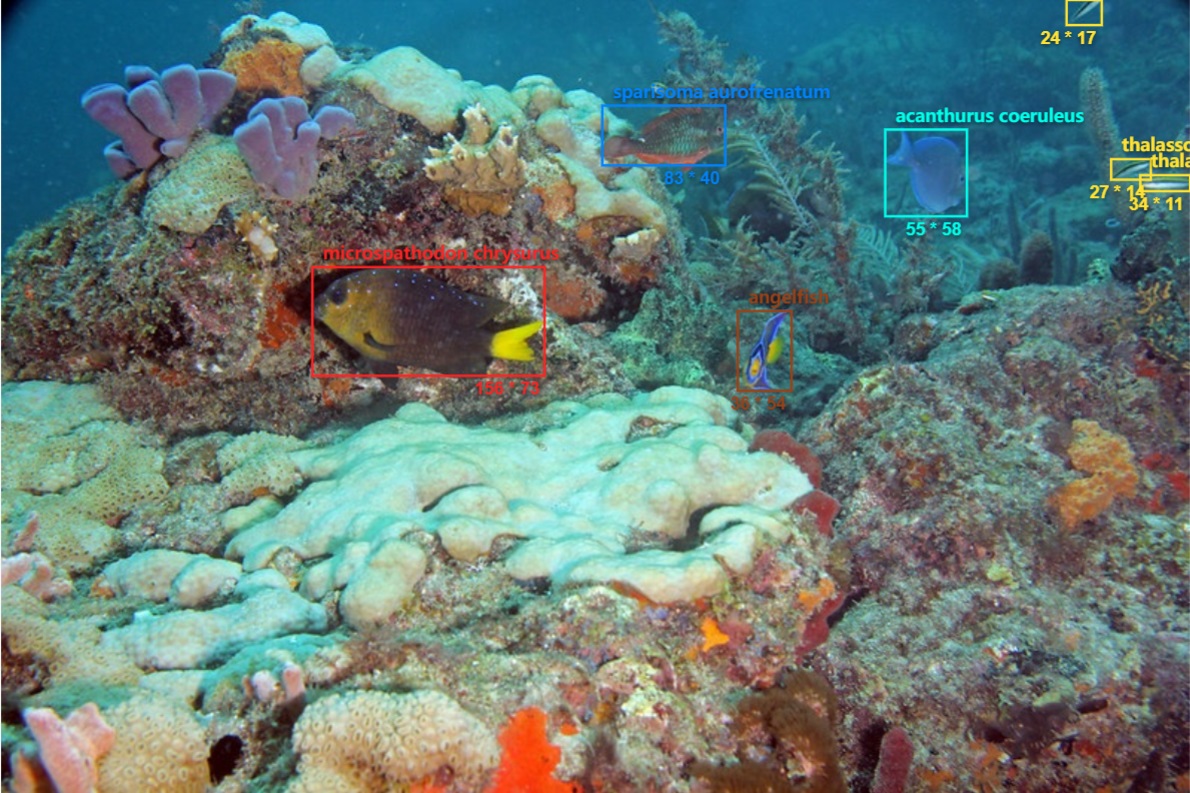} \\
    \end{tabular} \\
    \begin{tabular}{cccc}
        \includegraphics[height=3.7cm]{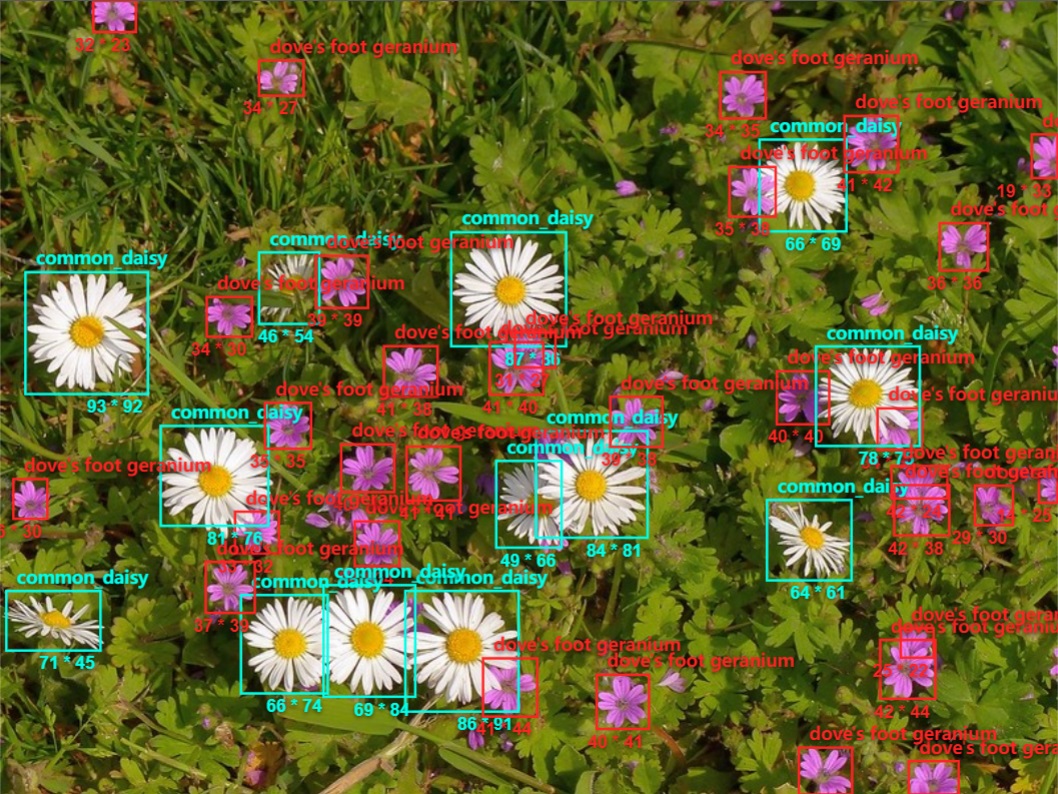} &
        \includegraphics[height=3.7cm]{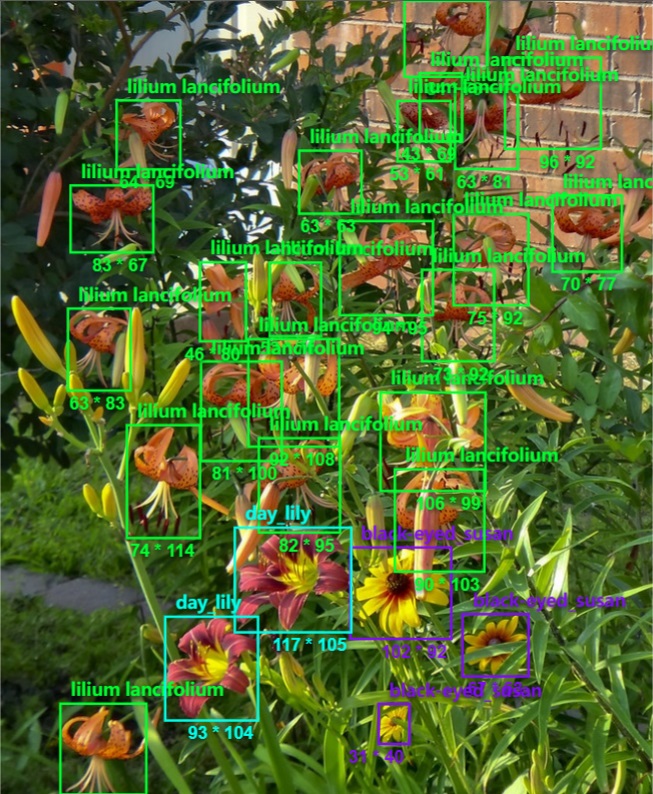} &
        \includegraphics[height=3.7cm]{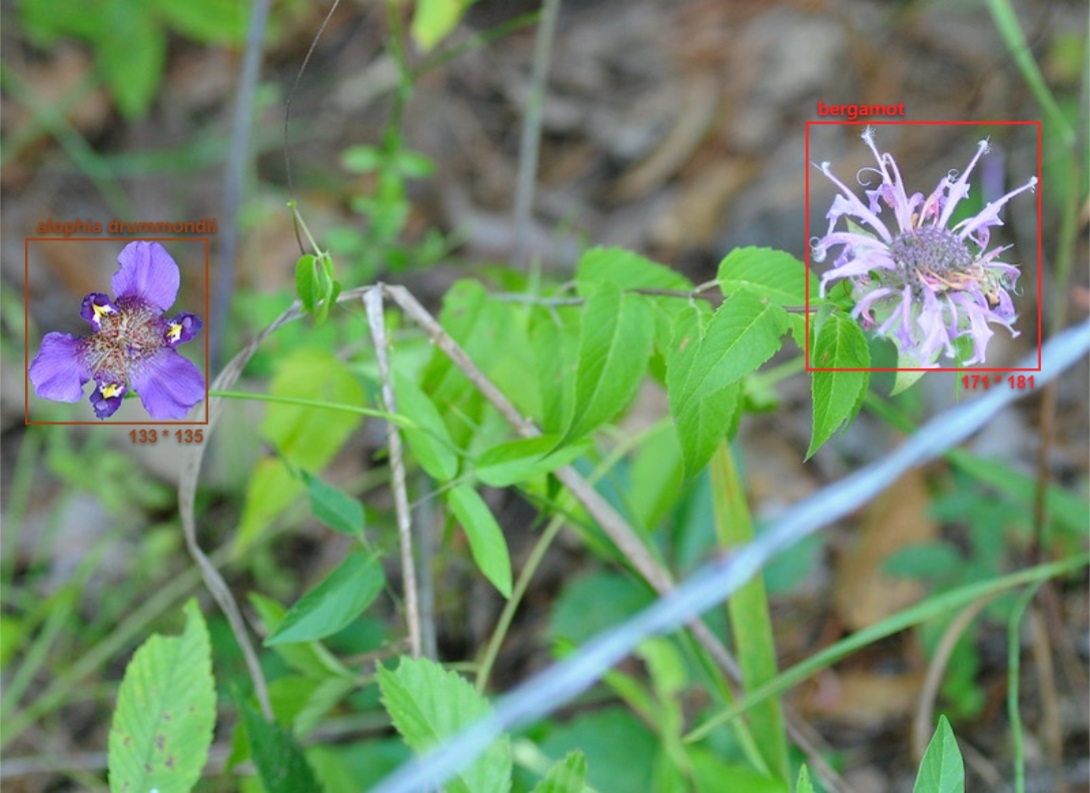} &
        \includegraphics[height=3.7cm]{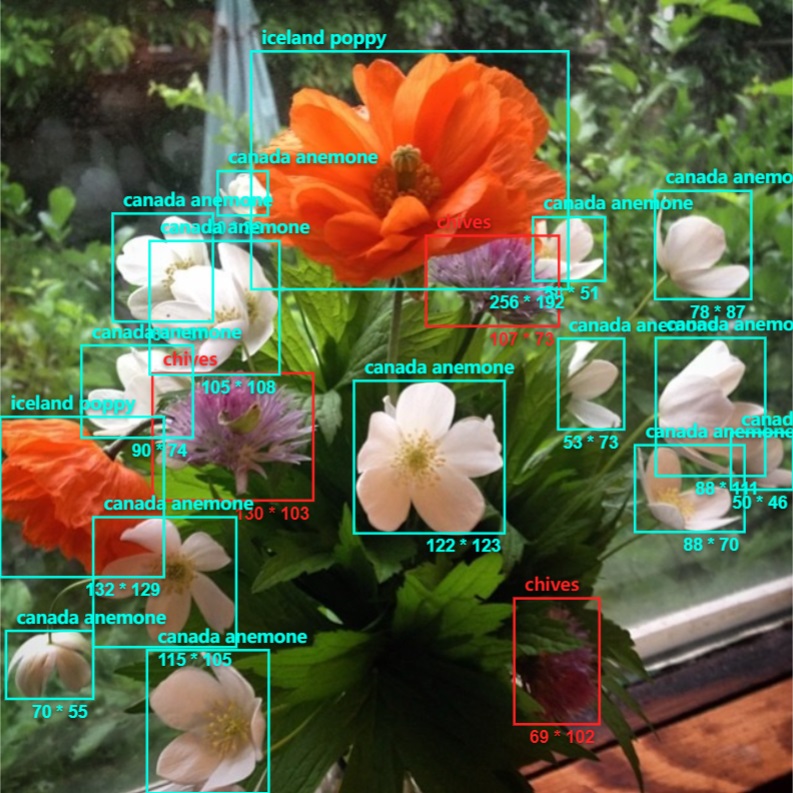} \\
    \end{tabular} \\
    \begin{tabular}{cccc}
        \includegraphics[height=2.98cm]{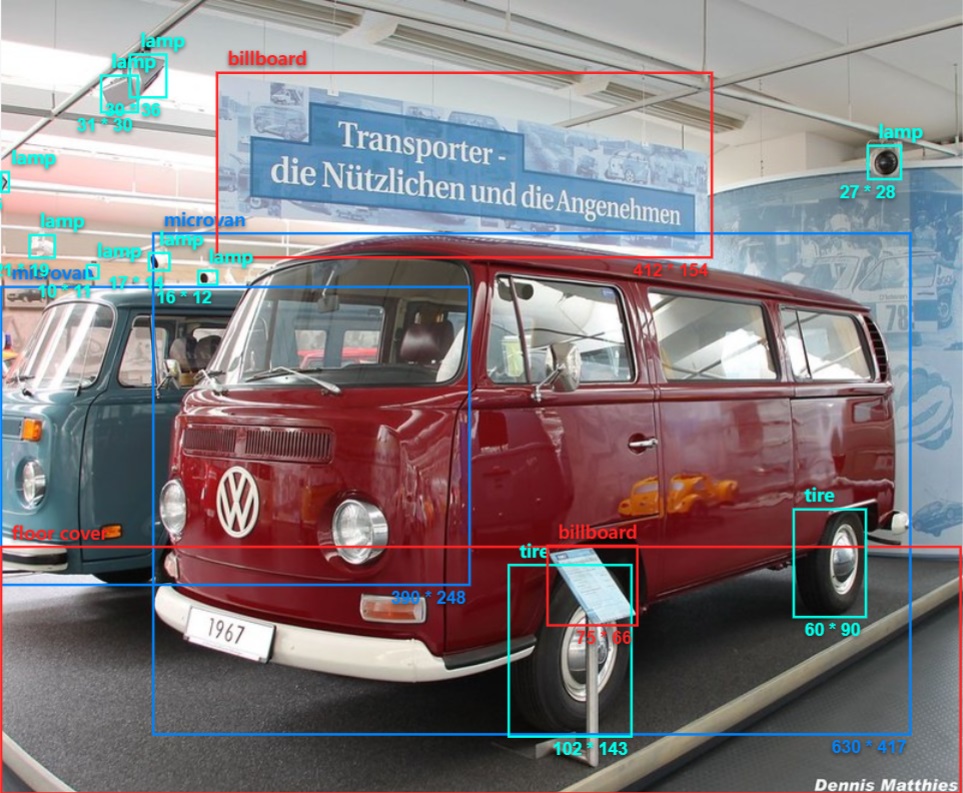} &
        \includegraphics[height=2.98cm]{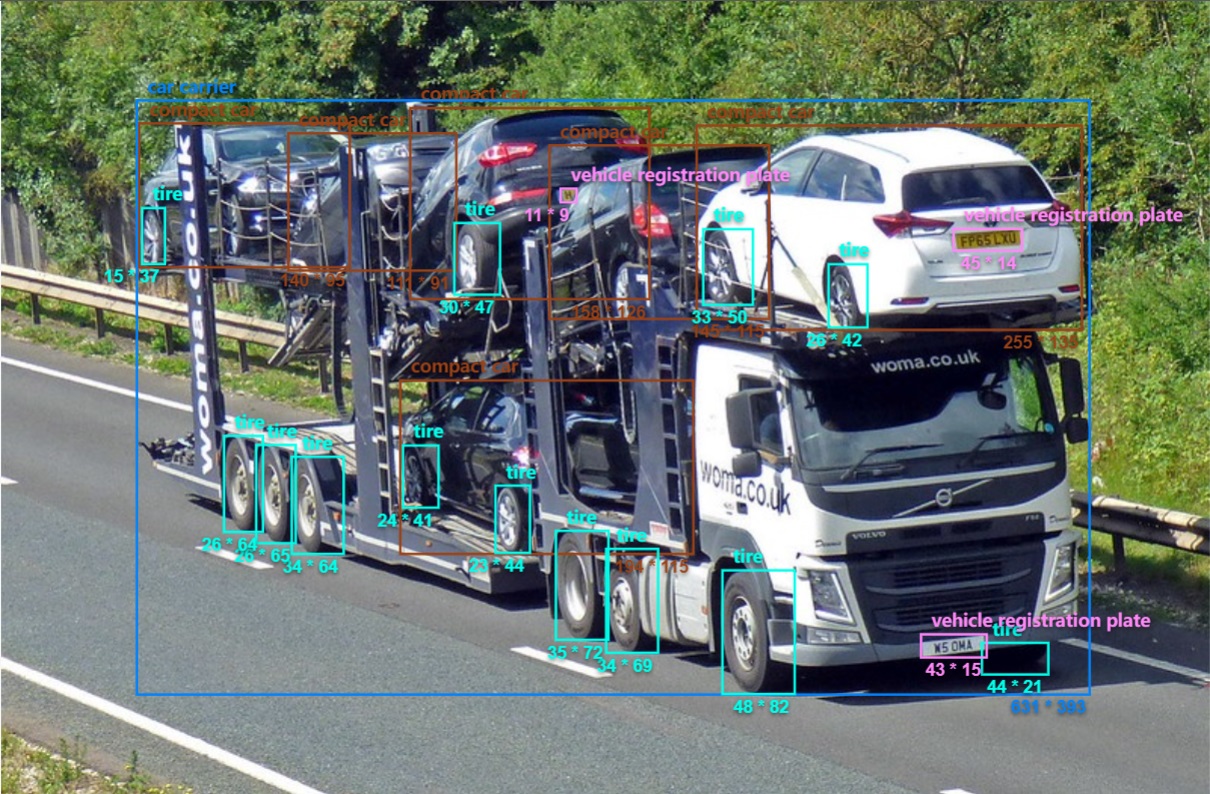} &
        \includegraphics[height=2.98cm]{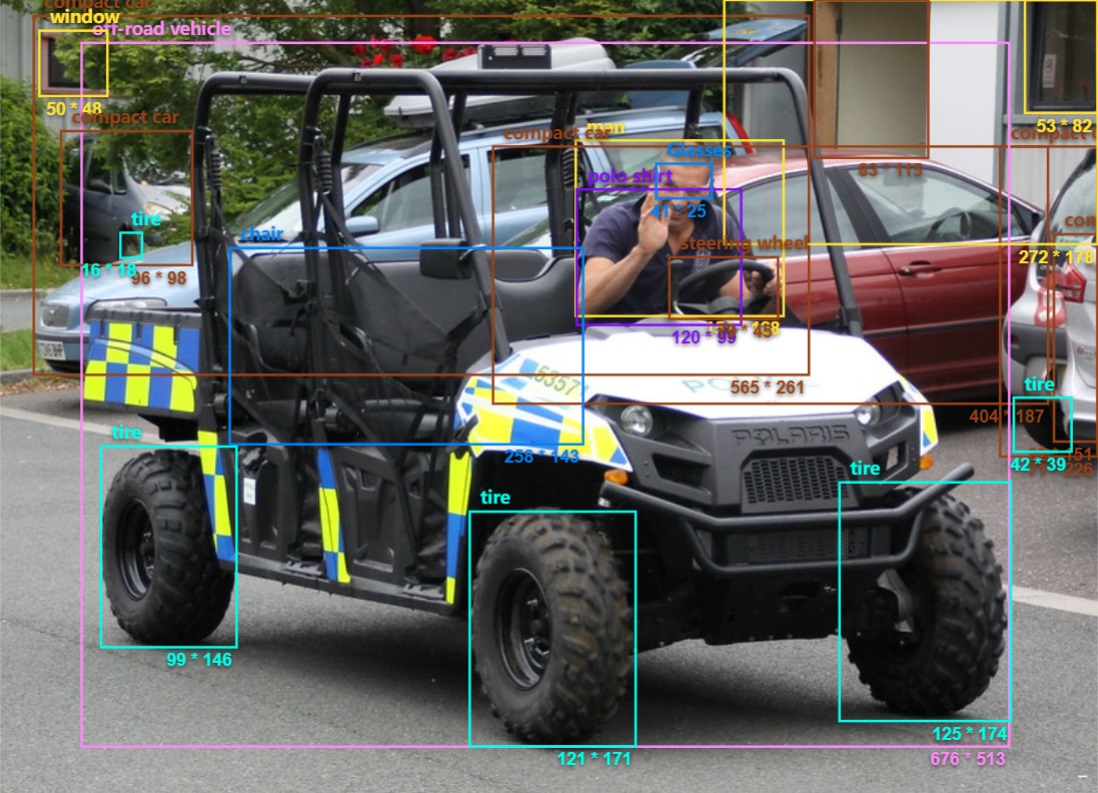} &
        \includegraphics[height=2.98cm]{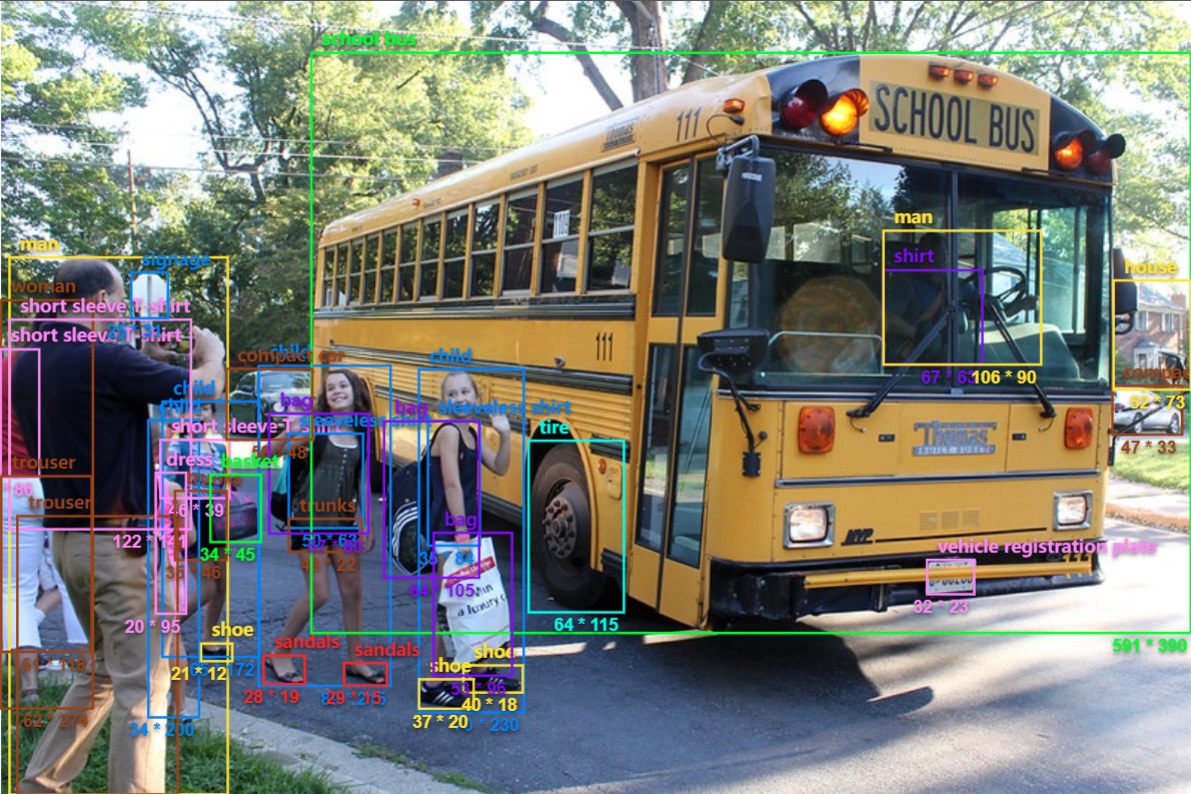} \\
    \end{tabular} \\
    \begin{tabular}{cccc}
        \includegraphics[height=2.955cm]{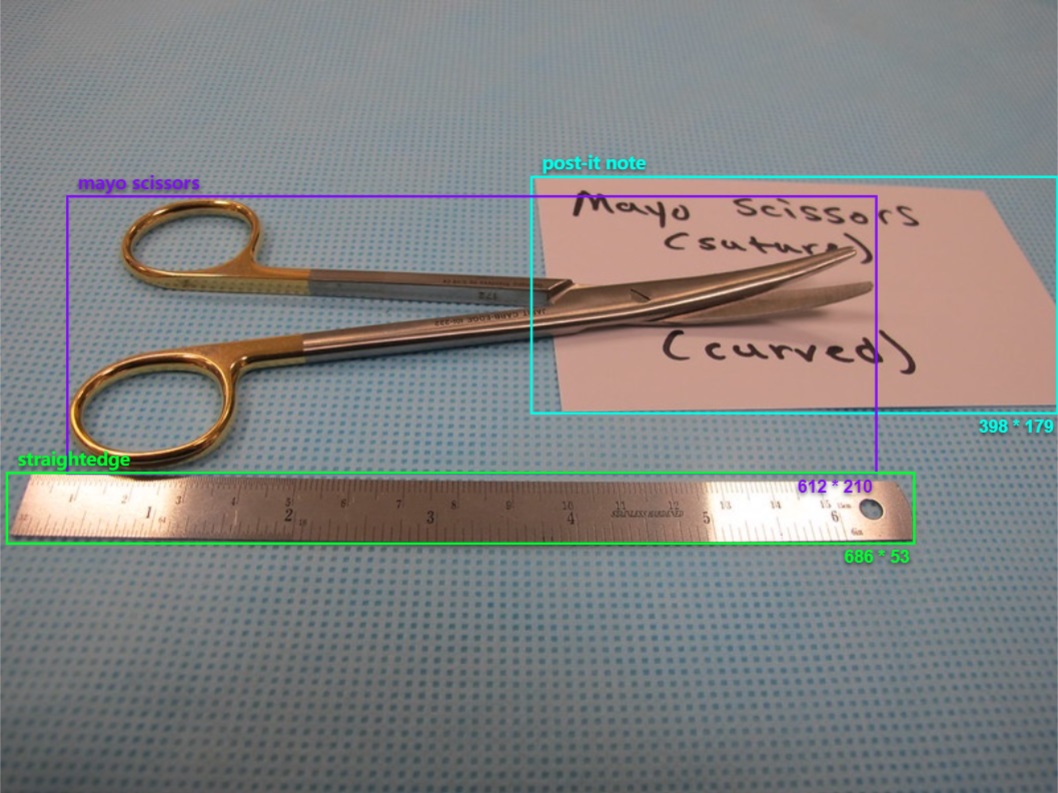} &
        \includegraphics[height=2.955cm]{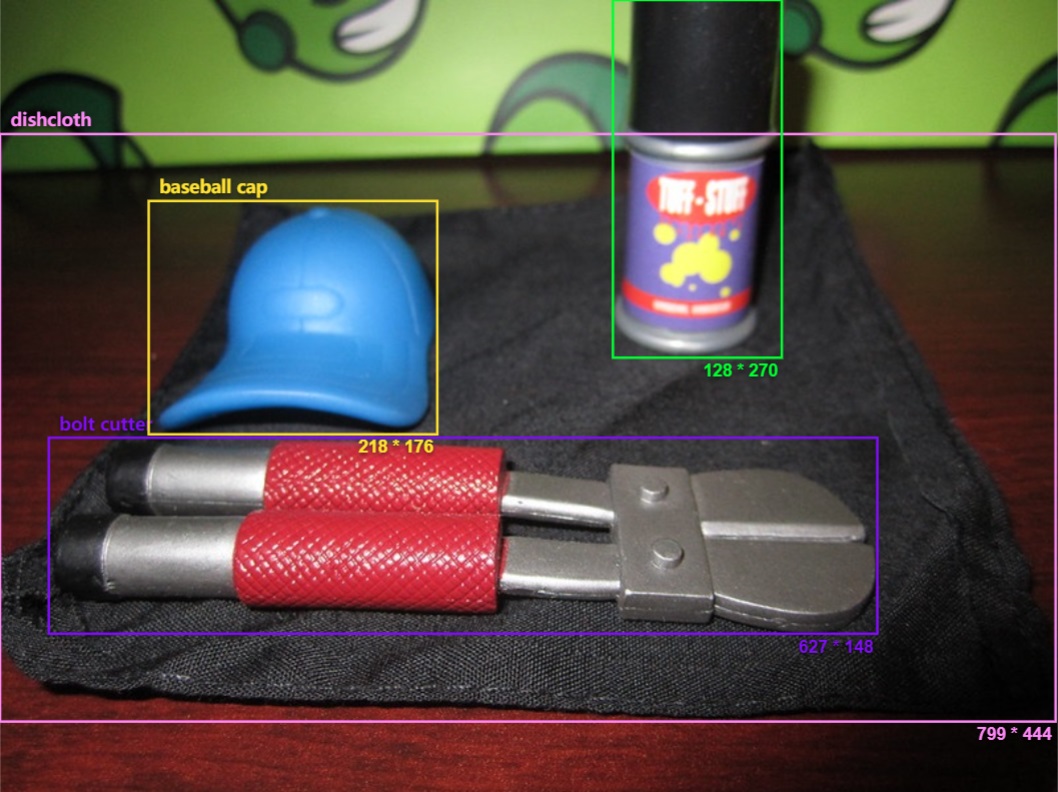} &
        \includegraphics[height=2.955cm]{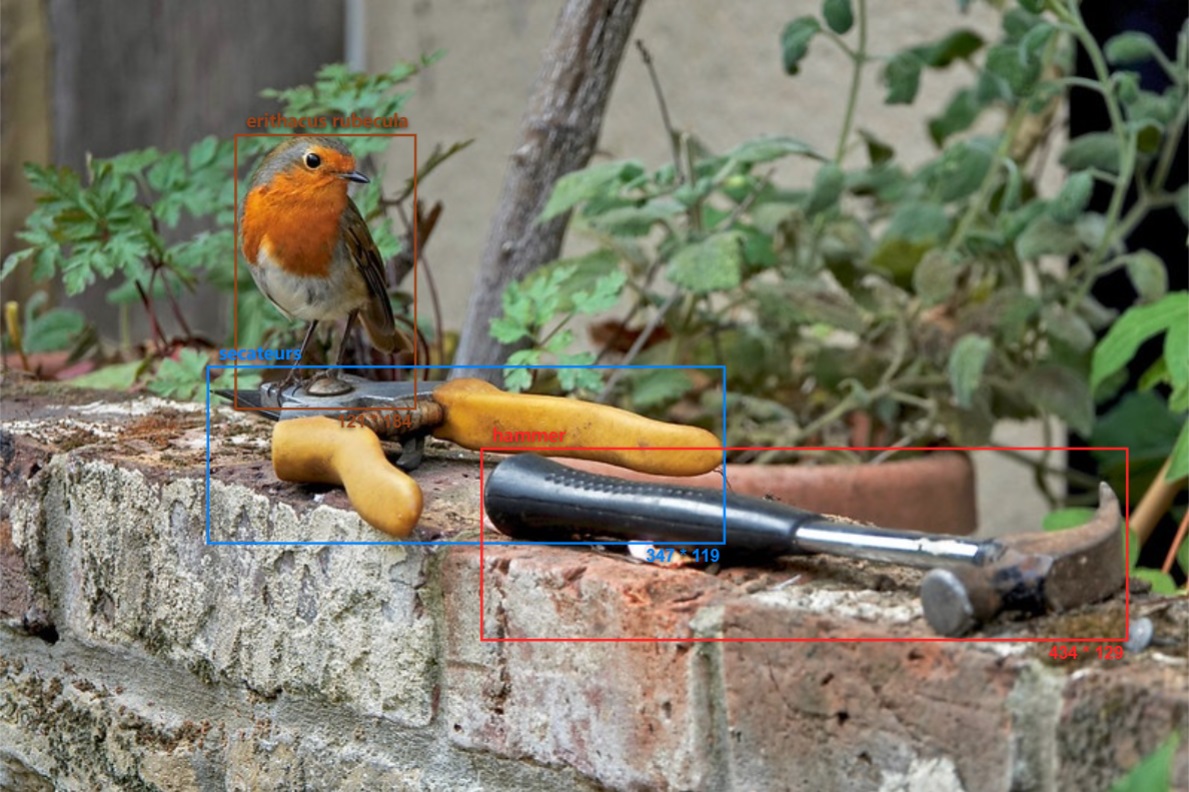} &
        \includegraphics[height=2.955cm]{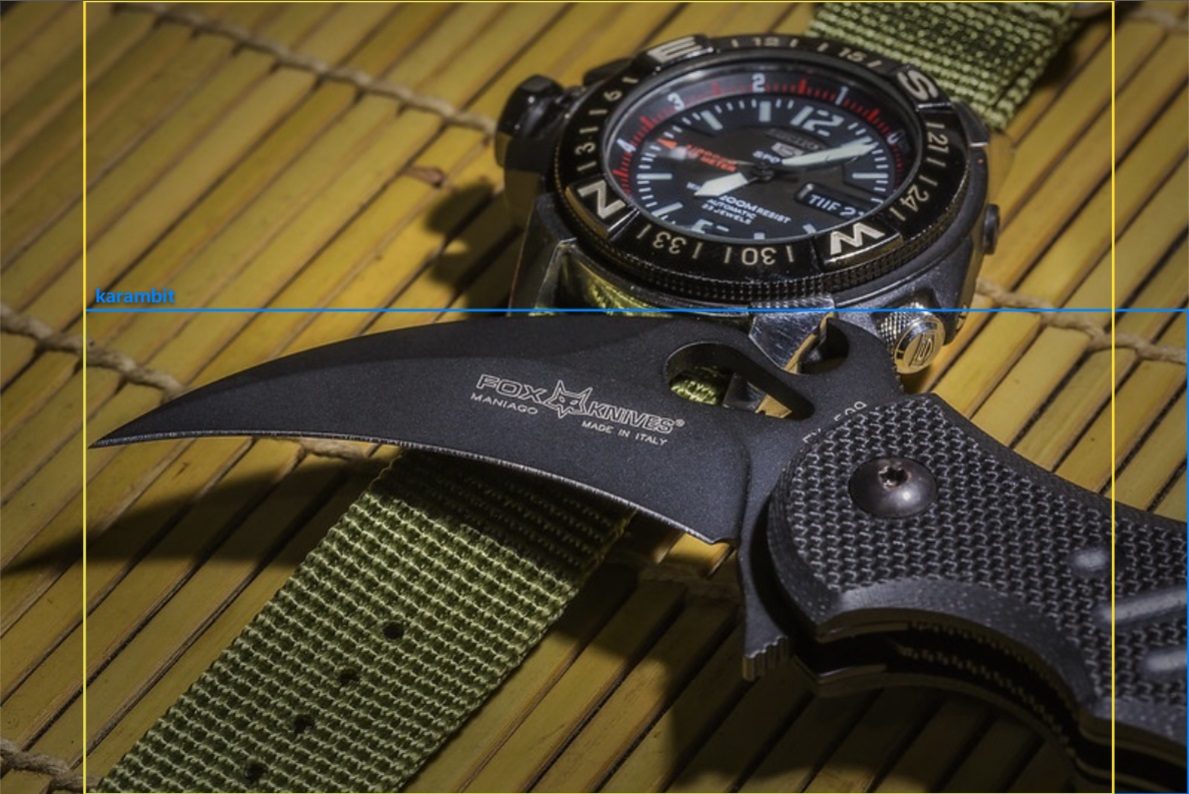}\\
    \end{tabular} \\
    }
    \vspace{-5pt}
\caption{\textbf{Visualizations of annotations in fine-grained categories.} Each row shows a set of visually similar category annotations that are easily confused.}
\label{figure:vis_simple}
\vspace{-7pt}
\end{figure*}

Figure~\ref{figure:vis_complex} and Figure~\ref{figure:vis_simple} provide some sampled images with annotations for visualization. 

\FloatBarrier

\FloatBarrier